\documentclass[lettersize,journal]{IEEEtran}
\usepackage{amsmath,amsfonts}
\usepackage{algorithmic}
\usepackage{algorithm}
\usepackage{array}
\usepackage[caption=false,font=scriptsize,labelfont=sf,textfont=sf]{subfig}
\usepackage{textcomp}
\usepackage{stfloats}
\usepackage{url}
\usepackage{verbatim}
\usepackage{graphicx}
\usepackage{cite}
\usepackage[justification=centering]{caption}
\usepackage{multirow}
\usepackage{bm}
\usepackage{booktabs}
\usepackage{color} 
\usepackage{makecell} 
\usepackage{orcidlink}
\hypersetup{hidelinks,
	colorlinks=true,
	allcolors=black,
	pdfstartview=Fit,
	breaklinks=true}
\begin{document}

\title{Zero-Shot Hashing Based on Reconstruction With Part Alignment}

\author{Yan Jiang$^{~\orcidlink{0009-0003-0820-1667}}$, Zhongmiao Qi, Jianhao Li, Jiangbo Qian$^{~\orcidlink{0000-0003-4245-3246}}$, ~\IEEEmembership{Member, IEEE}, Chong Wang \\and Yu Xin
        % <-this % stops a space

\vspace{-0.8cm}
\thanks{This work was supported in part by China NSF Grant No.62271274, Ningbo S\&T Project Grant No. 2024Z004, No. 2023Z059, and the programs sponsored by K. C. Wong Magna Fund in Ningbo University. \emph{(Corresponding author: Jiangbo Qian.)}}%

\thanks{Yan Jiang, Zhongmiao Qi, Jianhao Li, Jiangbo Qian, Chong Wang and Yu Xin are with the Faculty of Electrical Engineering and Computer Science, and Merchants' Guild Economics and Cultural Intelligent Computing Laboratory, Ningbo University, Ningbo 315211, China (e-mail: 2403567035@qq.com; 2111082371@nbu.edu.cn; 845156275@qq.com; qianjiangbo@nbu.edu.cn; wangchong@nbu.edu.cn; xinyu@nbu.edu.cn)}
}

\markboth{Journal of \LaTeX\ Class Files,~Vol.~14, No.~8, August~2021}%
{Shell \MakeLowercase{\textit{et al.}}: A Sample Article Using IEEEtran.cls for IEEE Journals}

\IEEEpubid{0000--0000/00\$00.00~\copyright~2021 IEEE}

\maketitle

\begin{abstract}
Hashing algorithms have been widely used in large-scale image retrieval tasks, especially for seen class data. Zero-shot hashing algorithms have been proposed to handle unseen class data. The key technique in these algorithms involves learning features from seen classes and transferring them to unseen classes, that is, aligning the feature embeddings between the seen and unseen classes. Most existing zero-shot hashing algorithms use the shared attributes between the two classes of interest to complete alignment tasks. However, the attributes are always described for a whole image, even though they represent specific parts of the image. Hence, these methods ignore the importance of aligning attributes with the corresponding image parts, which explicitly introduces noise and reduces the accuracy achieved when aligning the features of seen and unseen classes. To address this problem, we propose a new zero-shot hashing method called RAZH. We first use a clustering algorithm to group similar patches to image parts for attribute matching and then replace the image parts with the corresponding attribute vectors, gradually aligning each part with its nearest attribute. Extensive evaluation results demonstrate the superiority of the RAZH method over several state-of-the-art methods. 
\end{abstract}

\begin{IEEEkeywords}
Image retrieval, supervised hashing, zero-shot hashing, part alignment.
\end{IEEEkeywords}

\section{Introduction}
\IEEEPARstart{H}{ow} to efficiently process massive amounts of image data has garnered increasing attention in recent years. Hashing algorithms\cite{lu2016latent, cao2017hashnet} are widely used for image retrieval tasks because of their low storage and querying costs. Specifically, a hashing algorithm can encode high-dimensional data into compact binary codes and maintain the original data similarity through the Hamming distance measure.

\begin{figure}[!t]
	\centering
	\includegraphics{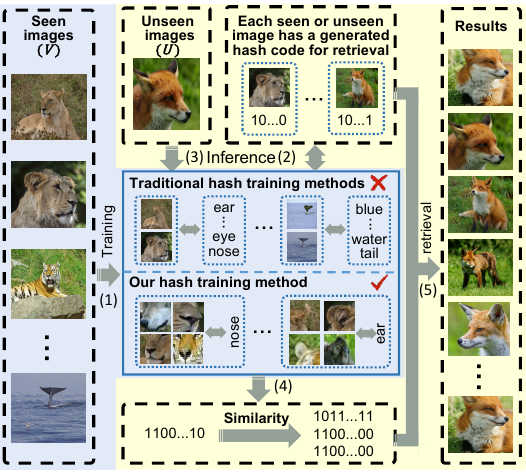}%
	\centering
	
	\captionsetup{justification=justified}
	\caption{Framework of zero-shot hashing algorithm. The blue background represents the model training process, while the yellow background represents the retrieval process.}
	\label{fig1}
	\vspace{-0.4cm}
\end{figure}

With their rapid development, deep neural networks have demonstrated outstanding performance in various tasks\cite{2018LSTMAutoencoders,2022imageto,2017tri,2019social,2023rethinking}, leading to the emergence of numerous deep hashing algorithms\cite{zhu2016deep, deng2019two,shen2021deep, shi2021transductive,wang2023deep}. The performance of deep neural networks heavily depends on the utilized training data. In the context of data explosion, however, unseen class data will inevitably appear during the retrieval process, and retraining the developed model would be costly. Therefore, it is better to learn a hashing model that can transfer seen classes to unseen classes. Driven by the concept of zero-shot learning\cite{hu2024domin, ma2024a,cheng2023discriminative}, zero-shot hashing methods\cite{zhang2019zero,lu2020adversarial,shi2022zero,xu2017attribute} are proposed. \hyperref[fig1]{Fig.1} illustrates the framework of the zero-shot hashing algorithm, where images are divided into two categories with no overlap: seen classes and unseen classes. During the training phase, only the seen classes are used. Once the training process is complete, the model parameters are frozen, and the images derived from the seen and unseen classes are input into the model to generate hash codes, enabling later retrieval tasks that are based on the similarity of these hash codes. As a result, the model is expected to generate highly accurate hash codes for new classes without retraining, allowing it to efficiently retrieve unseen images.

%\begin{figure}[!t]    % 常规操作\begin{figure}开头说明插入图片
%% 后面跟着的[htbp]是图片在文档中放置的位置，也称为浮动体的位置，关于这个我们后面的文章会聊聊，现在不管，照写就是了
%  \centering            % 前面说过，图片放置在中间
%  \subfloat[Universal]   % 第一张子图的下标（注意：注释要写在[]中括号内）
%  {
%      \label{fig:subfig1}\includegraphics[width=0.47\textwidth]{pics/fig1-1.pdf}
%  }\\
%  \subfloat[Ours]
%  {
%      \label{fig:subfig2}\includegraphics[width=0.47\textwidth]{pics/fig1-2.pdf}
%  }
%\captionsetup{font={small}, justification=raggedright}
%  \caption{Different zero-shot hashing alignment methods. }    % 整个图片的说明，注释写在{}内
%  \label{fig:subfig_1}            % 整个图片的标签编号，注意这里跟子图是一样的道理，标签不能重复 
%\end{figure}

\IEEEpubidadjcol Although the zero-shot hashing problem has attracted attention, the existing methods have the following weaknesses. These methods\cite{xu2017attribute, zhang2019zero} use convolutional neural networks(CNNs) to extract features from a whole image and then align them with attributes without considering the specific parts corresponding to those attributes. For example, as shown in Fig. 1, traditional methods always align the “ear” attribute with the whole lion image rather than the ear part. This approach ignores the importance of aligning attributes with their corresponding image parts, particularly the correspondence between attributes and the sizes of different image patches, which explicitly introduces noise and reduces the accuracy achieved during the alignment of seen and unseen class features. Therefore, as shown in Fig. 1, our method attempts to ensure high accuracy by aligning the “nose” attribute with the nose part. Here, we must solve two key problems: (1) how to capture relationships between image parts and attributes and (2) how the various sizes of image parts corresponding to attributes significantly increase the difficulty of matching.
 
Inspired by the excellent ViT method\cite{dosovitskiy2020image}, we propose a new zero-shot hashing method called RAZH (zero-shot hashing based on reconstruction with part alignment). RAZH first uses a clustering algorithm to group similar patches to image parts for attribute matching and then replaces the image parts with the corresponding attribute vectors, gradually aligning each part with its nearest attribute via a reconstruction strategy. 

%介绍零样本哈希算法，块和文字对应方法没有合适的，Transformer启发了我们，属性的对应有大小，因此在Transformer中引入聚类。

The main contributions of this paper can be summarized as follows.
\begin{itemize}
\item We propose that the key to solving the zero-shot hashing problem lies in achieving precise alignment between attributes and image parts. To address the challenge of aligning image parts of varying sizes with their corresponding attributes, we introduce a novel method called RAZH.
\item A two-branch reconstruction strategy, which combines a hash loss, a classification loss and a reconstruction loss, is proposed to optimize the model from both feature extraction and embedding alignment perspectives.
\item The experimental results show that the proposed RAZH method outperforms the state-of-the-art methods on zero-shot datasets.
\end{itemize}

The following sections of this paper are structured as follows. Section II outlines the previous related work. Section III details the RAZH method. Section IV presents the loss function used to optimize the model. Section V demonstrates the advancement exhibited by the RAZH method by conducting a comprehensive array of experiments. Finally, section VI provides a comprehensive summary of our work.

\section{RELATED WORK}
Zero-shot hashing methods have been developed from deep hashing methods and have produced many research results in recent years. RAZH differs from the previous methods in that it uses Transformer for feature extraction to complete the part alignment process, which provides a new potential solution for zero-shot hashing algorithms. This section briefly introduces the related work on zero-shot hashing, Transformer, and deep hashing.
\\ \hspace*{\fill} \\
\noindent \emph{A. Zero-shot Hashing Methods}

Because seen and unseen classes have different data distributions, zero-shot hashing can be used to transfer the supervision information derived from the seen data to the unseen data through the intermediate information. Xu et al.\cite{xu2017attribute} proposed attribute hashing (AH), which uses an MLP (multilayer perceptron) to implement the transfer of attribute information from seen to unseen classes. Inspired by unsupervised domain adaptation, Pachori et  al.\cite{pachori2018hashing} proposed ZSH-DA, which uses a max-margin classifier to connect visual features and semantic representations. To further ensure appropriate Hamming distances between different classes, Zhang et al.\cite{zhang2019zero} proposed a hashing method based on the orthogonal projection of images and semantic attributes; this approach enhances the discriminative nature of Hamming space delineation by requiring orthogonal relationships between the semantic attributes of images belonging to different classes. Zhong et al.\cite{zhong2020semantic} used class semantic embeddings as intermediate information to optimize the model via a combination of popular structure preservation and label constraints. Shi et al.\cite{shi2022zero} analyzed the impact of the similarity matrix on zero-shot hashing performance from another perspective and designed an asymmetric similarity matrix-constrained hash code, which can effectively alleviate the overfitting problem encountered when performing zero-shot hashing to seen classes. For multilabel zero-shot hashing, Zou et al.\cite{zou2020transductive} used the instance concept consistency ranking of the known class to anticipate labels for the unknown class and then employed the predicted labels as supervision information to guide the learning process of the hashing model. To address the problem of cross-modal hashing, Ji et al.\cite{ji2020attribute} proposed AgNet, which uses a high-dimensional semantic attribute space to align data across different modalities, combining attribute similarity with similar category regularization for a single modality. Shu et al.\cite{shu2022discrete} proposed DAZSH, which integrates pairwise similarity, class attributes, and semantic labels to guide the zero-shot hashing-based learning process. It combines data features with class attributes to capture the relationships between seen and unseen classes. 
\\ \hspace*{\fill} \\
\noindent \emph{B. Transformer Methods}

Transformer\cite{vaswani2017attention} is a type of neural network that employs a multiheaded attention mechanism to handle sequence-to-sequence tasks. Recently, Transformer methods, such as BERT and GPT, have become mainstream models in the field of natural language processing. In recent years, Transformer architecture has been successfully employed in computer vision tasks, resulting in better performance. For example, Dosovitskiy et al.\cite{dosovitskiy2020image} introduced the vision transformer (ViT), which segments a sample into multiple fixed-length patch sequences, uses Transformer to capture the global relationships among different regions within the input image, and incorporates the image information into a classification token for classification and prediction purposes. To solve the multiscale visualization problem encountered by Transformer, Liu et al.\cite{liu2021swin} proposed Swin Transformer, which implements a hierarchical structure using shifted windows, considers the flexibility of multiscale modeling and exhibits improved efficiency. By combining Transformer and object detection, Carion et al.\cite{carion2020end} proposed the DERT model, which learns relationships from the global context of an image by querying a fixed number of learnable objects and directly outputs the final result, eliminating the anchor generation step. Inspired by the self-supervised approach, which combines ViT and mask reconstruction strategies, He et al.\cite{he2022masked} proposed MAE, where random masked image patches are reconstructed via an asymmetric encoder-decoder, which enables the network to learn images with more discriminative row features.
\\ \hspace*{\fill} \\
\noindent \emph{C. Deep Hashing Methods}

\emph{1) Unsupervised Deep Hashing Methods: }Unsupervised hashing methods do not utilise supervised informationa and learn similarities from the given data distribution. Dong et al.\cite{dong2020unsupervised} proposed the unsupervised deep K-means hashing (UDKH) method, which uses the binary value obtained from clustering to perform supervision and gradually optimizes the clustering labels and hash codes by minimizing the pairwise supervisory loss to achieve effective image retrieval. To establish fine-grained similarity constraints in unsupervised deep hashing algorithms, Qin et al.\cite{qin2021unsupervised} proposed an unsupervised deep multisimilarity hashing method with a semantic structure (UDMSH), which constructs a similarity matrix with a semantic structure by calculating the cosine distances between the deep features of image pairs to guide the process of learning hash codes. To ensure that appropriate interhash code distances are generated by the unsupervised method, Ma et al.\cite{ma2019toward} proposed distinguishable unsupervised graph hashing (DUGH), which constructs a KNN graph by utilizing a random projection tree and employing a neighborhood search technique, effectively accounting for both the similarities and differences contained within the initial dataset. Cao et al.\cite{cao2024unsupervised} integrated contrastive learning into the unsupervised hashing method and proposed a fine-grained similarity-preserving contrastive learning-based hashing method that enhances the distinguishability of hash codes by reconstructing the fine-grained similarities. To solve the unsupervised hashing problem encountered in cases with multilabel data, Shi et al.\cite{shi2023unsupervised} proposed a new unsupervised adaptive feature selection model with binary hashing (UAFS-BH), which uses spectral embedding and image content constraints to adaptively determine the number of image pseudolabels needed to guide the network training process. Zhang et al.\cite{zhang2021aggregation} used aggregated graph convolutional hashing (AGCH) to solve the cross-modal unsupervised hashing problem; AGCH uses mode-specific encoders, graph convolutional networks and fusion modules to learn from the joint similarity matrix constructed by combining the structural information of multiple modes and can effectively mine the correlations between different modes.

\emph{2) Supervised Deep Hashing Methods:} Supervised deep hashing methods leverage supervisory label information to empower the constructed network to generate more discriminative hash codes. Lu et al.\cite{lu2020adversarial} proposed an adversarial multilabel variational hashing (AMVH) method that uses generators and discriminators for hash code learning, fitting the network to unseen data during training. To address the feature information loss induced by quantization strategies and discrete optimization, Li et al.\cite{li2022deep} proposed a novel deep attention-guided hashing method with pairwise labels (DAHP), which enhances the global feature fusion and contextual information learning process by introducing an anchor hash code generation algorithm and utilizing positional and channel attention mechanisms. To learn hash codes with multiple lengths at the same time, Nie et al.\cite{nie2022supervised} proposed supervised discrete multilength hashing (SDMLH), which can obtain hash codes with arbitrary lengths by fusing the information possessed by hash codes with different lengths. To fully exploit the supervised information contained in multilabel data, Shi et al.\cite{shi2022supervised} proposed SASH, which uses a similarity matrix as its optimization objective to determine the relevance of labels by maintaining the spatial consistency of the features and labels, based on the spatial consistency of the feature labels. To consider the connections between different objects in a sample, Peng et al.\cite{peng2023multi} designed the multilabel hashing with multiobject dependency (DRMH) method, which uses an object detection network to extract the features of an object and then fuses the acquired global features and location features to establish dependencies between objects. With the emergence of the ViT, SWTH\cite{peng2023swin} was proposed as a hashing model with a Swin transformer as its backbone network, which can effectively obtain hash codes containing the overall context of the input sample. To solve the parameter optimization problem caused by ignoring the distribution of the given samples, Qin et al.\cite{qin2023deep} proposed deep adaptive quadruplet hashing with probability sampling (DAQH), which enhances the information possessed by the training batch.
\\ \hspace*{\fill} \\
\noindent \emph{D. Summary}

The existing zero-shot hashing methods improve the retrieval performance achieved for unseen classes by both leveraging the supervisory information derived from seen classes and designing effective loss functions. However, as they employ CNNs for image feature extraction and directly map the entire input image to an attribute embedding, they disregard the inherent associations between attributes and specific parts of an image. Consequently, these methods result in suboptimal hash codes for unseen classes.

\begin{table}[!t]
\centering
\caption{}{DESCRIPTION OF SYMBOLS}
\\ [0.2cm]
\setlength{\tabcolsep}{2.0mm}{
\renewcommand{\arraystretch}{1.1}
\begin{tabular}{cc}
\midrule
{Symbol} & {Description}                                                  \\ \hline
$\bm{x}_i$             & Input sample $i$                                                              \\
$\bm{b}_i$            & { Hash codes related to input sample $i$} \\
$\bm{y}_i$              & {Label vector related to input sample $i$}          \\
$\bm{C}$              & {Classes of samples}          \\
$\bm{V}$              & {Seen classes of samples}          \\
$\bm{U}$              & {Unseen classes of samples}          \\
$\bm{T}$              & {Attributes of all classes}          \\
$N$              		&{Amount of samples}             \\
$M$              		&{Amount of patches in a sample                                 }             \\
$K$              		&{Length of hash codes }             \\
 \midrule
\end{tabular}}
\vspace{-0.4cm}
\end{table}

\begin{figure*}[!t]
	\centering
	\includegraphics[width=7.1in]{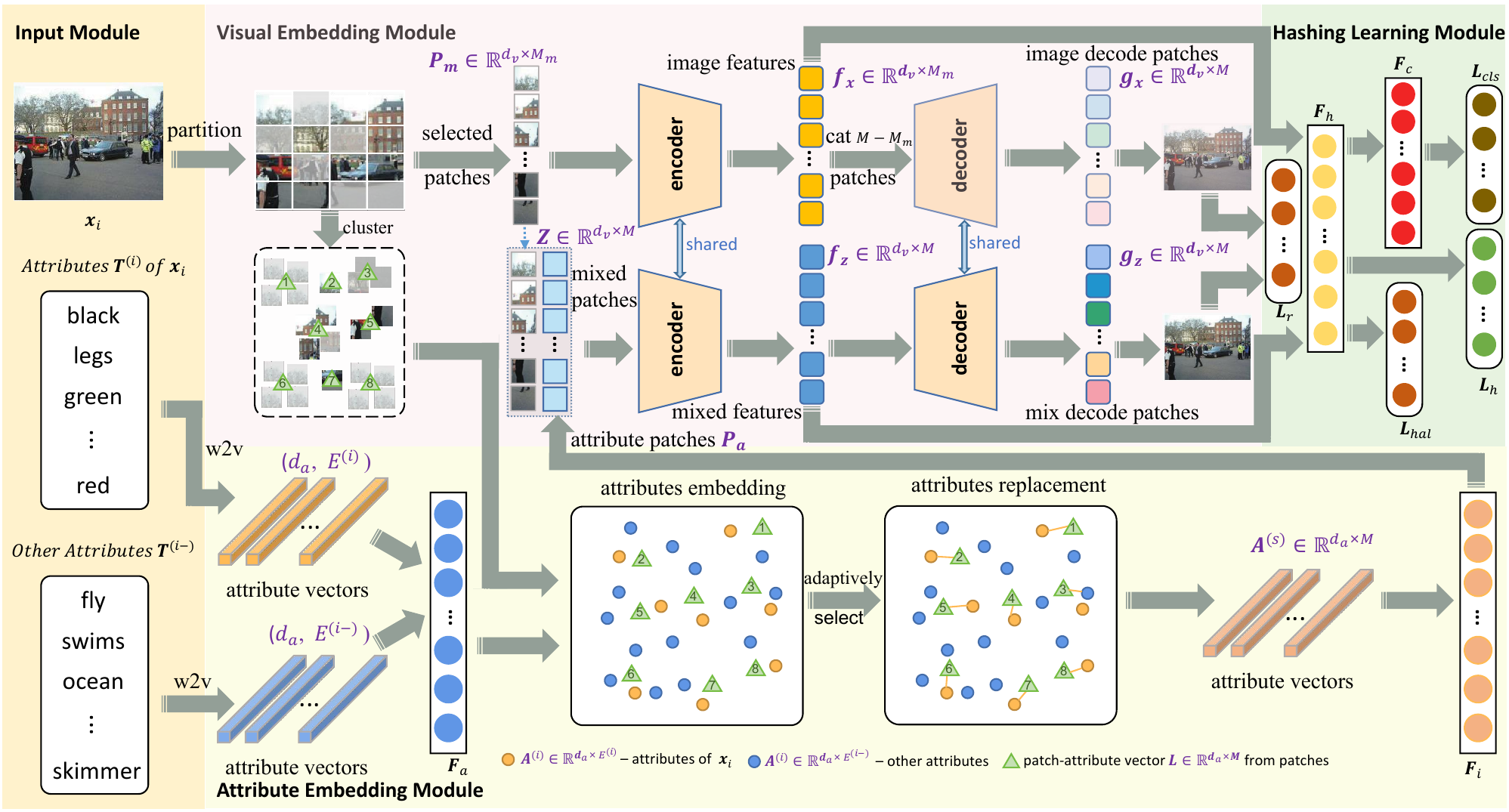}%
	\centering
	\captionsetup{justification=justified}
	\caption{The architecture of RAZH, which is composed of four key modules: 1) an input module: inputs to the model for training; 2) a visual embedding module: aligns patches and attributes through reconstruction operations; 3) an attribute embedding module: replaces patches in attribute embedding; and 4) a hashing learning module: integrates overall losses to train the model. The idea behind RAZH is to replace unselected image patches with attribute patches that have identical semantic information, thereby fusing the image and attribute data for joint training.}
	\label{fig2}

\end{figure*}

\section{PROPOSED RAZH METHOD}

 $\boldsymbol{X}=\left\{\bm{x}_{1}, \cdots \cdots, \boldsymbol{x}_{N}\right\}$ denotes the $N$ samples contained in the training set, and the class label matrix is defined as $\boldsymbol{Y}=\left\{\boldsymbol{y}_{1}, \cdots \cdots, \boldsymbol{y}_{N}\right\} $. In the zero-hashing setting, all classes are divided into seen classes $\bm{V}$ and unseen classes $\bm{U}$, where $\bm{V}\cup \bm{U}= \bm{C}, \bm{V}\cap \bm{U}= \emptyset$. The attributes shared by all classes are $\boldsymbol{T}=\left\{\bm{t}_{1}, \cdots \cdots, \boldsymbol{t}_{N^{'}}\right\}$, where $N^{'}$ denotes the number of attributes. For each $\bm{x}_i$, there is an attribute set $\bm{T}^{(i)}$ that describes $\bm{x}_i$ and another attribute set $\bm{T}^{(i-)}$, where $\bm{T}^{(i)}\cup \bm{T}^{(i-)}= \bm{T}$ and $\bm{T}^i\cap \bm{T}^{(i-)}= \emptyset$. The goal of zero-shot hashing is to learn the mapping function  $\mathcal{F}: \boldsymbol{X} \rightarrow \boldsymbol{B} \in\{-1,+1\}^{K \times N}$ through the seen classes, which maps all classes to Hamming space, that is, converts $\bm{x}_i$ into a compact $K$-bit binary code. In Table \uppercase\expandafter{\romannumeral1}, we present a summary of the notations used.
\\ \hspace*{\fill} \\
\noindent \emph{A.	Network Architecture}

To address the problem mentioned above, the model needs to learn the parts of the input image that correspond to various attributes. The idea behind RAZH is to replace an image part with an attribute that has identical semantic information, thereby fusing the image and attribute data for joint training. To achieve this, RAZH first divides the given image into patches and employs clustering and reconstruction strategies to progressively align these parts with the attributes. In this way, RAZH is able to transfer supervisory information from the seen classes to the unseen classes, thereby generating reliable hash codes. \hyperref[fig2]{Fig.2} illustrates the architecture of RAZH, which consists of four main modules: an input module, a visual embedding module, an attribute embedding module, and a hash learning module.
\\ \hspace*{\fill} \\
\noindent \emph{B.	Detailed Module Introduction}

\emph{1) Input Module}: The input module consists of samples $\bm{x}_i$ and attributes. The attributes are divided into two parts: the attributes $\bm{T}^{(i)}$ of $\bm{x}_i$ and the other attributes $\bm{T}^{(i-)}$.

\emph{2) Visual Embedding Module}: Since directly segmenting images based on attributes is challenging, we first divide an image into fixed-sized patches. As shown in \hyperref[fig2]{Fig.2}, the image $\bm{x}_i$ is first fed into the visual embedding module and reshaped into a sequence of flattened patches. Each patch is then linearly projected onto a $d_v$-dimensional space, resulting in a patch set $\bm{P}\in \mathbb{R}^{d_v\times M}$, where $M$ denotes the total number of patches.

We then introduce a novel dual-branch contrastive reconstruction structure, a mixing branch and a reconstruction branch, enabling not only effective image feature extraction but also the precise alignment of image patches with their corresponding attributes. The mixing branch enhances the degree of semantic alignment by replacing image patches with matching attributes and then reconstructing the original image, effectively guiding the encoder to align the semantic information of different image parts with their corresponding attributes. The reconstruction branch reconstructs the initial image by utilizing selected patches, enabling the encoder to learn a better feature representation. This dual-branch approach strengthens the ability of the encoder to capture global structure and local attribute details, producing a more balanced feature representation. 

\noindent $\bullet$ \textbf{ Mixing Branch}

For the mixing branch, the main goal is to replace image patches with their corresponding attributes, enabling semantic alignment through reconstruction. However, a fixed image patch size  may inadequately cover attribute regions of various sizes, potentially affecting the accuracy of the attribute representations. To address this issue, RAZH employs K-means clustering to group patches with similar features, dynamically adjusting the segmentation granularity level. This ensures that patches corresponding to the same attribute are effectively clustered, enhancing the degree of semantic coherence and improving the attribute representations. That is,
\begin{equation}
	\boldsymbol{j}_{k}=\frac{1}{\left|C_{k}\right|} \sum_{\boldsymbol{p}_{i} \in C_{k}} \boldsymbol{p}_{i}, \quad \forall i \in\{1,2, \ldots, K^{'}\}
\end{equation}
where $\bm{j}_k$ is the center of the \(k\)-th cluster, $K^{'}$ is the number of cluster, \(|C_k|\) is the number of patches contained in the \(k\)-th cluster and \(\bm{p}_i\) is a patch belonging to the \(k\)-th cluster.

The cluster centers are then used as representative features for the image attributes, enabling the measurement of the similarities between the clusters and attribute data. To achieve this, we use a linear layer $\bm{F}_m$ to map \(\bm{J} = \{\bm{j}_1, \bm{j}_2, \dots, \bm{j}_{K^{'}}\}\) to \(\bm{L}\), and the mapping is then fed into an attribute embedding module (AEM), which adaptively identifies the semantically nearest attribute for each image part. The AEM subsequently provides the attribute patches Pa that semantically align most closely with \(\bm{L}\). The details of the AEM are presented later.
\begin{equation}
	\begin{array}{c}\boldsymbol{L}=\boldsymbol{F}_{\boldsymbol{m}}\left(\boldsymbol{J}\right), \\\boldsymbol{P}_{a}=\operatorname{AEM}\left(\boldsymbol{L}\right),\end{array}
\end{equation}

To minimize the noise derived from incorrect matches, we set a threshold and replace image parts with attribute vectors only if their similarity exceeds the set threshold. That is,
\begin{equation}
	\boldsymbol{Z}=\left\{\begin{array}{ll}\boldsymbol{P}_{a}, & \text { sim } \geq \text { threthold } \\\boldsymbol{P}, & \text { sim }<\text { threctrold }\end{array}\right.
\end{equation}
where \(\text{sim}\) represents the similarity between an image patch and the corresponding attribute patch. The obtained $\bm{Z}$ is subsequently input into the encoder to obtain an encoded feature $\bm{f}_z\in \mathbb{R}^{d_v\times M}$. Finally, $\bm{f}_z$ is passed through the decoder to generate the reconstructed result of the mixed patches, $\bm{g}_z\in \mathbb{R}^{d_v\times M}$. That is,
\begin{equation}
	\begin{aligned}\boldsymbol{f}_{\boldsymbol{z}} & =\text { Encoder }(\boldsymbol{Z}), \\\boldsymbol{g}_{\boldsymbol{z}} & =\text { Decoder }\left(\boldsymbol{f}_{\boldsymbol{z}}\right).\end{aligned}
\end{equation}

\noindent $\bullet$ \textbf{ Reconstruction Branch}

For the reconstruction branch, the main goal is to enhance the feature extraction ability of the encoder. To reduce the computational complexity of the model and extract more representative features, a subset of \(M_m\) patches is randomly selected from $\boldsymbol{P}$, and these patches are referred to as the selected image patches $\bm{P}_m\in \mathbb{R}^{d_v\times M_m}$. $\bm{P}_m$ is utilized for extracting image features, which are directly fed into the encoder to obtain the encoded feature $\bm{f}_x\in \mathbb{R}^{d_v\times M_m}$. To ensure that the decoder receives comprehensive information about the entire image, we concatenate the shared and learnable vectors of the \(M-M_m\) patches with $\bm{f}_x$, resulting in a combined feature set \(\bm{f}_x^{'} \in \mathbb{R}^{d_v \times M}\). This concatenated feature set $\bm{f}_x^{'}$ is subsequently fed into the shared decoder to reconstruct the image. That is,
\begin{equation}
	\begin{aligned}\boldsymbol{f}_{\boldsymbol{x}} & =\text { Encoder }(\boldsymbol{P}_m), \\\boldsymbol{g}_{\boldsymbol{x}} & =\text { Decoder }(\boldsymbol{f}_{\boldsymbol{x}}^{'}).\end{aligned}
\end{equation}

\emph{3) Attribute Embedding Module (AEM)}: For each cluster center, the attribute embedding module (AEM) adaptively searches for the nearest attribute and assigns it to the associated patch. This ensures that image patches are aligned with their corresponding attributes during the joint training process, thereby improving features extracted for zero-shot hashing.

First, the attributes $\bm{T}^{(i)}$ and $\bm{T}^{(i-)}$ are separately fed into the AEM and transformed into word vectors via the glove method (w2v)\cite{pennington2014glove}. Consequently, a fully connected layer $\bm{F}_a$ is employed to map these word vectors into two attribute subspaces, $\bm{A}^{(i)}\in \mathbb{R}^{d_a\times E^{(i)}}$  and $\bm{A}^{(i-)}\in \mathbb{R}^{d_a\times E^{(i-)}}$; one is the attribute embedding of $\bm{x}_i$, and the other is the other attribute embedding. Here, $E^{(i)}$ denotes the number of attributes belonging to $\bm{x}_i$, and $E^{(i-)}$ denotes the number of other attributes. The similarity between $\bm{L}$ and $\bm{A}^{(i)}$ is then quantified via the cosine similarity measure, and the attribute vector $\bm{A}^{(s)}$ with the highest similarity is selected, where $\bm{L}$ represents the image patch mapped to the attribute embedding space. Finally, the output of the attribute embedding module is obtained by increasing the dimensionality of $\bm{A}^{(s)}$ to $d_v$ through the fully connected layer $\bm{F}_i$. That is,
\begin{equation}
\begin{array}{c}\boldsymbol{A}^{(i)}, \boldsymbol{A}^{(i-)}=\boldsymbol{F}_{a}\left(w 2 v\left(\boldsymbol{T}^{(i)}, \boldsymbol{T}^{(i-)}\right)\right), \\\boldsymbol{A}^{(s)}=\operatorname{Top}\_\operatorname{sim}\left(\boldsymbol{A}^{(i)}, \boldsymbol{L}\right), \\\boldsymbol{P}_{\boldsymbol{a}}=\boldsymbol{F}_{i}\left(\boldsymbol{A}^{(s)}\right),\end{array}
\end{equation}
where $\operatorname{Top}\_\operatorname{sim}(\cdot)$ denotes a function that selects attribute vectors based on highest similarity values and $\bm{P}_a$ denotes the semantic attribute patches.

\emph{4) Hash Learning Module}: The hash learning module is employed to generate hash codes for retrieval purposes. When a sample $\bm{x}_i$ is input into the network, we acquire the corresponding sample feature $\bm{f}_x\in \mathbb{R}^{d_v}$ through the encoder and then perform hash learning based on the basis of this sample feature. Specifically, the hashing layer is defined as follows: 
\begin{equation}
\boldsymbol{h}_{i}=\tanh \left(\boldsymbol{W}^{T}_{h} \boldsymbol{f}_{x}+\boldsymbol{o}_{h}\right),
\end{equation}
where $\bm{W}_{h}\in \mathbb{R}^{d_v\times K}$ and $\bm{o}_{h}\in \mathbb{R}^K$ denote the weight and bias of the hash layer, correspondingly, and $tanh(\tau) = (e^{\tau} -e^{-\tau})/(e^{\tau}+e^{-\tau})$ is used to smooth continuous values. Then, the binarization function $\operatorname{sgn(\cdot)}$ is used to generate the hash code used for retrieval:
\begin{equation}
\boldsymbol{b}_{i}=\operatorname{sgn}\left(\boldsymbol{h}_{i}\right),
\end{equation}
where $\bm{b}_i$ is the hash code generated by the sample $\bm{x}_i$.

\section{OVERALL LOSS}
To generate hash codes that better match images, we design an overall loss function that includes a hash loss, a cross-entropy loss, and a reconstruction loss. These three losses are integrated with different weights to optimize the model, which can effectively learn information from the seen classes.
\\ \hspace*{\fill} \\
\noindent \emph{A.	Hash Loss}

In hash loss, we construct the similarity matrix $\bm{S}=\{s_{ij}\}_{i,j=1}^{N}\in \{0,1\}^{N\times N}$ based on the class labels corresponding to the samples. If $\bm{x}_i$ and $\bm{x}_j$ have at least one identical class, then $s_{ij} = 1$; otherwise, $s_{ij} = 0$. Given a minibatch of hash codes, we transform the hash constraint problem into a probability problem under specific conditions:
\begin{equation}
\begin{split}p\left(s_{i j} \mid \boldsymbol{b}_{i}, \boldsymbol{b}_{j}\right) &=\left\{\begin{array}{r}\psi\left(\delta_{i j}\right) , \quad s_{i j}=1 \\1-\psi\left(\delta_{i j}\right), \quad s_{i j}=0\end{array} \right. \\&=\psi\left(\delta_{i j}\right)^{s_{i j}}\left(1-\psi\left(\delta_{i j}\right)\right)^{1-s_{i j}},\end{split}
\end{equation}
where $\psi(\tau)=1/(1+e^{-{\tau}})$ denotes the sigmoid function and $\delta_{ij}=\frac{1}{2}\left \langle \boldsymbol{b}_i,\boldsymbol{b}_j\right \rangle$.

For example, when $\bm{b}_i$ and $\bm{b}_j$ are far apart, the inner product $\left \langle \boldsymbol{b}_i, \boldsymbol{b}_j\right \rangle$ is smaller and $p(0 | \boldsymbol{b}_i,\boldsymbol{b}_j)$ is large, implying that $\bm{x}_i$ and $\bm{x}_j$ have been classified as dissimilar, and that the hash loss is smaller. In other words, a smaller value of $p(s_{ij} | b_i,b_j)$ value should result in a larger loss and vice versa.

We utilize weighted negative log-likelihood values to measure the similarity constraints between samples, and use it as the loss for backpropagation, which is defined as follows:
\begin{equation}
\begin{split}\mathcal{L}_{h}&=-\sum_{i=1}^{N} \sum_{j=1}^{N} \theta_{i j} \log \left(p\left(s_{i j} \mid \boldsymbol{b}_{i}, \boldsymbol{b}_{j}\right)\right) \\&=\sum_{i=1}^{N} \sum_{j=1}^{N} \theta_{i j}\left(\log \left(1+e^{\frac{1}{2} \boldsymbol{b}_{i}^{T} \boldsymbol{b}_{j}}\right)-\frac{1}{2} s_{i j} \boldsymbol{b}_{i}^{T} \boldsymbol{b}_{j}\right),\end{split}
\end{equation}
where $\theta_{ij}$ is the weight set used to handle imbalances on $\bm{S}$; this parameter is set to:
\begin{equation}
\theta_{i j}=\left\{\begin{array}{ll}N / s_{p}, & s_{i j}=1 \\N / s_{n}, & s_{i j}=0\end{array},\right.
\end{equation}
where $s_{p}=\frac{1}{2} \sum_{i=1}^{N} \sum_{j=1}^{N} s_{i j}$  is the number of similar samples and $s_{n}=N-\frac{1}{2} \sum_{i=1}^{N} \sum_{j=1}^{N} s_{i j}$  is the number of dissimilar samples. The hash loss enhances the similarity of the hash codes generated by similar samples and increases the hash code distances between dissimilar samples.
\\ \hspace*{\fill} \\
\noindent \emph{B. Classification Loss}

To extract more discriminative features, we introduce a classification loss,which is defined as follows:
\begin{equation}
\mathcal{L}_{c l s}=-\sum_{i=1}^{N} \boldsymbol{y}_{i} \log \left(\tilde{\boldsymbol{y}}_{i}\right),
\end{equation}
where $\boldsymbol{y}_{i}$ represents the label vector corresponding to sample $\bm{x}_i$ and $\tilde{\boldsymbol{y}}_{i}$  is the output generated by the classification layer, which is defined as $\tilde{\boldsymbol{y}}_{i}=\operatorname{Softmax}\left(\tilde{\boldsymbol{W}}^{T} \boldsymbol{h}_{i}+\tilde{\boldsymbol{o}}\right)$,  $\operatorname{Softmax}(\boldsymbol{\tau})=e^{\boldsymbol{\tau}} / \sum_{i=1}^{C} e^{\boldsymbol{\tau}_{i}}$. $\tilde{\boldsymbol{W}} \in \mathbb{R}^{K \times C}$  and $\tilde{\boldsymbol{o}} \in \mathbb{R}^{C}$ denote the weights and biases of the classification layers, respectively.
\\ \hspace*{\fill} \\
\noindent \emph{C. Reconstruction Loss}

The method proposed in this paper has two reconstruction tasks, and we use the loss settings of MAE and MSE to measure the reconstruction ability of the model to reconstruct the input image. Only the loss of the selected patch is calculated as follows:
\begin{equation}
\mathcal{L}_{r}=\sum_{i=1}^{N}\left(\boldsymbol{g}_{z}-\boldsymbol{P}\right)^{2}+\left(\boldsymbol{g}_{x}-\boldsymbol{P}\right)^{2},
\end{equation}
where $\boldsymbol{g}_{z}$ and $\boldsymbol{g}_{x}$ denote the reconstruction of selected patches and the reconstruction of mixed patches derived from $\bm{x}_i$, respectively.

To make the original image and the mixed image closer at the hash code level, a hash code alignment loss is constructed. That is:
\begin{equation}
\mathcal{L}_{ {hal }}=\frac{1}{2 N} \sum_{i=1}^{N}\left|\boldsymbol{b}_{i}^{a}-\boldsymbol{b}_{i}\right|,
\end{equation}
where $\bm{b}_i^a$ denotes the hash code of the mixed image and $\bm{b}_i$ denotes the hash code of the original image. Therefore, the overall reconstruction loss is defined as shown below:
\begin{equation}
\mathcal{L}_{ {rec }}=\mathcal{L}_{ {r }}+\mathcal{L}_{ {hal }}.
\end{equation}
\\ \hspace*{\fill} \\
\noindent \emph{D. Optimization}

Combining the above losses, the total losses used for optimization can be expressed as follows:
\begin{equation}
\mathcal{L}=\mathcal{L}_{{cls }}+\alpha\mathcal{L}_{{h }}+\beta\mathcal{L}_{{rec }},
\end{equation}
where $\alpha$ and $\beta$ are the balancing factors of the loss function. According to Eq. (16), generating binary hash codes ensures that similar samples generate similar hash codes and that dissimilar samples have larger Hamming distances.

In the training phase, we use the back-propagation algorithm and gradient descent to optimize the loss function. Owing to the non-differentiable nature of the absolute value function $\left| \cdot \right| $ and the complexity involved in computing its derivative, we use a smoothing alternative, replacing $\left| b \right|$ with $log(cosh(b))$\cite{hyvarinen2009natural}. It is difficult to integrate binarization into the training process, so the output of the hashing layer $\bm{h}_i$ is used instead. Therefore, Eq.(10) can be redefined as follows:

\begin{equation}
\begin{split}\mathcal{L}_{h}&=-\sum_{i=1}^{N} \sum_{j=1}^{N} \theta_{i j} \log \left(p\left(s_{i j} \mid \boldsymbol{h}_{i}, \boldsymbol{h}_{j}\right)\right) \\&=\sum_{i=1}^{N} \sum_{j=1}^{N} \theta_{i j}\left(\log \left(1+e^{\frac{1}{2} \boldsymbol{h}_{i}^{T} \boldsymbol{h}_{j}}\right)-\frac{1}{2} s_{i j} \boldsymbol{h}_{i}^{T} \boldsymbol{h}_{j}\right),\end{split}
\end{equation}
and Eq.(14) can be redefined as shown below:

\begin{equation}
\mathcal{L}_{{hal }}=\frac{1}{2 N} \sum_{i=1}^{N} \log \left(\cosh \left(\boldsymbol{h}_{i}^{a}-\boldsymbol{h}_{i}\right)\right).
\end{equation}

For Eq.(17), the $\mathcal{L}_h$ gradient of the output $\bm{h}_i$ relative to the hashing layer can be expressed as $\frac{\partial {\mathcal{L}_h}}{\partial \bm{h}_i}$, which is calculated as follows:
\begin{equation}
\frac{\partial {\mathcal{L}_h}}{\partial \bm{h}_i} = \sum_{i=1}^N\sum_{i=1}^N\frac{\theta_{ij}}{2}\times \frac{\bm{h}_je^{{\frac{1}{2}\bm{h}^T_i\bm{h}_j}}}{1+e^{{\frac{1}{2}\bm{h}^T_i\bm{h}_j}}}-\frac{\theta_{ij}}{2}s_{ij}\bm{h}_j.
\end{equation}

For Eq.(12), the $\mathcal{L}_{cls}$ gradient of the output $\stackrel{\sim}{\bm{y}}_{i}$ relative to the classification layer can be expressed as  $ \frac{\partial {\mathcal{L}_{cls}}}{\partial \stackrel{\sim}{\bm{y}}_{i}} $, which is calculated as follows:
\begin{equation}
\frac{\partial {\mathcal{L}_{cls}}}{\partial \stackrel{\sim}{\bm{y}}_{i}} = -\sum_{i=1}^N \frac{\bm{y}_i}{\stackrel{\sim}{\bm{y}}_{i}}.
\end{equation}

For Eq.(15), the $\mathcal{L}_{rec}$ gradient of the output $\bm{g}_x$ relative to the reconstruction layer can be expressed as  $ \frac{\partial {\mathcal{L}_{rec}}}{\partial \bm{g}_x}$, which is calculated as follows:
\begin{equation}
\frac{\partial \mathcal{L}_{rec}}{\partial \boldsymbol{g}_{x}}=2\sum_{i=1}^{N} \left(\boldsymbol{g}_{x}-\boldsymbol{P}\right).
\end{equation}

For Eq.(15), the $\mathcal{L}_{rec}$ gradient of the output $\bm{h}_i$ relative to the hashing layer output can be expressed as  $ \frac{\partial {\mathcal{L}_{rec}}}{\partial \bm{h}_i}$, which is calculated as follows:
\begin{equation}
\frac{\partial \mathcal{L}_{r e c}}{\partial \boldsymbol{h}_{i}}=\frac{1}{2 N} \sum_{i=1}^{N} \tanh \left(\boldsymbol{h}_{i}^{a}-\boldsymbol{h}_{i}\right).
\end{equation}

We employ a backpropagation algorithm utilizing gradient information, which leverages the chain rule to optimize the overall loss. Algorithm 1 shows the detailed process of RAZH.
\vspace{-0.4cm}
\begin{algorithm}
	\renewcommand{\algorithmicrequire}{\textbf{Input:}}
	\renewcommand{\algorithmicensure}{\textbf{Output:}}
	\caption{RAZH}
	\label{alg:1}
	\begin{algorithmic}
		\REQUIRE Image data of training data $\bm{X}$, Label data of training data $\bm{Y}$, Attributes of class $\bm{T}$, Hash code length $K$; Maximum number of epochs $Max\_Epochs$;  Balancing factors $\alpha$, $\beta$
		\ENSURE Hash codes $\bm{B}$ and model;
		\FOR{$epoch$=0, 1, ..., $Max\_Epochs$}
		\STATE{To construct a batch from image data $\bm{X}$}
		 \FOR{$\bm{x}$ in batch}
		\STATE{Get patches $\bm{P}$ by linearly project and perform patch clustering on $\bm{P}$ using K-means;}
		\STATE{Get arribute patches $\bm{P_a}$ that semantically exhibit the nearest attribute by attribute embedding module;}
		\STATE{Get reconstruction patches $\bm{g_z}$ and $\bm{g_x}$ by visual embedding module;}
		\STATE{Acquire the output $\bm{h}$ of the hash layer;}
		\STATE{Acquire the output $\tilde{\bm{y}}$ of classification layer;}
		 \ENDFOR
		\STATE{Compute the total loss for each sample according to equation (16);}
		\STATE{Compute gradients according to Eq.(19), (20), (21) and (22);}
		\STATE{Optimization of parameters by back-propagation;}
		\ENDFOR
		\STATE \textbf{return} hash codes $\bm{B}$ and model.
	\end{algorithmic} 
	 
\end{algorithm}
\vspace{-0.4cm}

\section{EXPERIMENT}
We perform comprehensive experiments on the extensively employed benchmark datasets CIFAR10\cite{krizhevsky2009learning}, CUB\cite{wah2011caltech} and AWA2\cite{xian2018zero} datasets to verify the performance of RAZH.

The experiments in this paper are conducted using a training set and test set following the requirements of zero-shot hashing. All dataset settings in this paper follow \cite{shi2022zero, peng2023multi, jiang2018asymmetric} etc.
\\ \hspace*{\fill} \\
\noindent \emph{A. Datasets}

\textbf{CIFAR10} consists of 60,000 hand-annotated images belonging to 10 classes, with 6,000 images per class. In the experiments, 8 classes are considered seen classes, and the remaining classes are considered unseen classes. Under the experimental settings, the training set contains 4,000 images, with each class consisting of a set of 500 randomly chosen images; the testing set contains 200 images, with each class consisting of a set of 100 randomly chosen images; and the remaining 55,800 images are included in the retrieval set.

\begin{table*}[!t]
\centering
\caption{}{RESULTS OF mAP@5000(\%) FOR VARIOUS CODE LENGTHS ACROSS THREE DATASETS ARE PRESENTED. THE FIRST AND SECOND HALVES DISPLAY THE mAP OUTCOMES ACHIEVED BY OUR APPROACH UTILIZING LINEAR HASHING AND DEEP HASHING, RESPECTIVELY.}
\label{tab2}
\\ [0.2cm]
\setlength{\tabcolsep}{2.2mm}{
\renewcommand{\arraystretch}{1.5}
\begin{tabular}{ccccccccccccc}
\hline
\multicolumn{1}{c|}{\multirow{2}{*}{\textbf{Method}}} & \multicolumn{4}{c|}{\textbf{CIFAR10}}                                                                                                                        & \multicolumn{4}{c|}{\textbf{CUB}}                                                                                                                         & \multicolumn{4}{c}{\textbf{AWA2}}                                                                                                 \\ \cline{2-13} 
\multicolumn{1}{c|}{}                                 & \multicolumn{1}{c|}{24bits}          & \multicolumn{1}{c|}{48bits}          & \multicolumn{1}{c|}{64bits}          & \multicolumn{1}{c|}{128bits}         & \multicolumn{1}{c|}{24bits}          & \multicolumn{1}{c|}{48bits}          & \multicolumn{1}{c|}{64bits}          & \multicolumn{1}{c|}{128bits}         & \multicolumn{1}{c|}{24bits}          & \multicolumn{1}{c|}{48bits}          & \multicolumn{1}{c|}{64bits}          & 128bits         \\ \hline

\multicolumn{1}{c|}{LSH}                                                       & \multicolumn{1}{c|}{0.89}          & \multicolumn{1}{c|}{1.51}          & \multicolumn{1}{c|}{2.02}          & \multicolumn{1}{c|}{2.08}          & \multicolumn{1}{c|}{0.55}          & \multicolumn{1}{c|}{0.69}          & \multicolumn{1}{c|}{0.76}          & \multicolumn{1}{c|}{0.95}          & \multicolumn{1}{c|}{1.06}  & \multicolumn{1}{c|}{1.51}  & \multicolumn{1}{c|}{2.04}  & 3.06          \\
\multicolumn{1}{c|}{SH}                                                        & \multicolumn{1}{c|}{7.26}          & \multicolumn{1}{c|}{8.50}          & \multicolumn{1}{c|}{8.56}          & \multicolumn{1}{c|}{8.81}          & \multicolumn{1}{c|}{5.68}          & \multicolumn{1}{c|}{8.10}          & \multicolumn{1}{c|}{8.86}          & \multicolumn{1}{c|}{11.91}          & \multicolumn{1}{c|}{18.33}  & \multicolumn{1}{c|}{27.29}  & \multicolumn{1}{c|}{29.55}  & 34.41          \\
\multicolumn{1}{c|}{ITQ}                                                       & \multicolumn{1}{c|}{9.29}          & \multicolumn{1}{c|}{8.78}          & \multicolumn{1}{c|}{9.36}          & \multicolumn{1}{c|}{10.57}          & \multicolumn{1}{c|}{5.33}          & \multicolumn{1}{c|}{7.65}          & \multicolumn{1}{c|}{8.92}          & \multicolumn{1}{c|}{11.82}          & \multicolumn{1}{c|}{19.99}  & \multicolumn{1}{c|}{28.21}  & \multicolumn{1}{c|}{29.64}  & 37.64          \\
\multicolumn{1}{c|}{SDH}                                                       & \multicolumn{1}{c|}{7.96}          & \multicolumn{1}{c|}{8.72}          & \multicolumn{1}{c|}{9.13}          & \multicolumn{1}{c|}{10.07}          & \multicolumn{1}{c|}{4.95}          & \multicolumn{1}{c|}{8.74}          & \multicolumn{1}{c|}{10.69}          & \multicolumn{1}{c|}{10.28}          & \multicolumn{1}{c|}{14.60}  & \multicolumn{1}{c|}{24.81}  & \multicolumn{1}{c|}{29.75}  & 33.26          \\
\multicolumn{1}{c|}{FSSH}                                                    & \multicolumn{1}{c|}{8.99}          & \multicolumn{1}{c|}{8.49}          & \multicolumn{1}{c|}{11.55}          & \multicolumn{1}{c|}{12.75}          & \multicolumn{1}{c|}{0.55}          & \multicolumn{1}{c|}{1.31}          & \multicolumn{1}{c|}{1.66}          & \multicolumn{1}{c|}{7.49}          & \multicolumn{1}{c|}{0.54}  & \multicolumn{1}{c|}{17.49}  & \multicolumn{1}{c|}{20.74}  & 29.37          \\
\multicolumn{1}{c|}{SCDH}                                                      & \multicolumn{1}{c|}{6.54}          & \multicolumn{1}{c|}{8.53}          & \multicolumn{1}{c|}{9.28}          & \multicolumn{1}{c|}{8.74}          & \multicolumn{1}{c|}{0.92}          & \multicolumn{1}{c|}{2.63}          & \multicolumn{1}{c|}{4.10}          & \multicolumn{1}{c|}{7.09}          & \multicolumn{1}{c|}{16.03}  & \multicolumn{1}{c|}{24.23}  & \multicolumn{1}{c|}{31.83}  & 28.80          \\
\multicolumn{1}{c|}{TSK}                                                      & \multicolumn{1}{c|}{9.39}          & \multicolumn{1}{c|}{10.93}          & \multicolumn{1}{c|}{11.44}          & \multicolumn{1}{c|}{12.79}          & \multicolumn{1}{c|}{7.39}          & \multicolumn{1}{c|}{12.00}          & \multicolumn{1}{c|}{13.94}          & \multicolumn{1}{c|}{11.12}          & \multicolumn{1}{c|}{22.62}  & \multicolumn{1}{c|}{31.09}  & \multicolumn{1}{c|}{38.73}  & 41.51          \\
\multicolumn{1}{c|}{AH}                                                        & \multicolumn{1}{c|}{8.99}          & \multicolumn{1}{c|}{10.49}          & \multicolumn{1}{c|}{11.08}          & \multicolumn{1}{c|}{13.53}          & \multicolumn{1}{c|}{4.80}          & \multicolumn{1}{c|}{8.97}          & \multicolumn{1}{c|}{10.89}          & \multicolumn{1}{c|}{14.45}          & \multicolumn{1}{c|}{22.75}  & \multicolumn{1}{c|}{19.89}  & \multicolumn{1}{c|}{31.54}  & 35.57          \\
\multicolumn{1}{c|}{OPZH}                                                      & \multicolumn{1}{c|}{9.11}          & \multicolumn{1}{c|}{10.63}          & \multicolumn{1}{c|}{10.72}          & \multicolumn{1}{c|}{12.76}          & \multicolumn{1}{c|}{6.3}          & \multicolumn{1}{c|}{8.79}          & \multicolumn{1}{c|}{9.62}          & \multicolumn{1}{c|}{11.43}          & \multicolumn{1}{c|}{10.56}  & \multicolumn{1}{c|}{13.90}  & \multicolumn{1}{c|}{16.18}  & 19.61          \\
\multicolumn{1}{c|}{ASZH}                                                      & \multicolumn{1}{c|}{11.03}           & \multicolumn{1}{c|}{14.71}          & \multicolumn{1}{c|}{15.68}          & \multicolumn{1}{c|}{16.37}          & \multicolumn{1}{c|}{7.64}          & \multicolumn{1}{c|}{11.92}          & \multicolumn{1}{c|}{12.94}          & \multicolumn{1}{c|}{17.27}          & \multicolumn{1}{c|}{26.19}  & \multicolumn{1}{c|}{37.87}  & \multicolumn{1}{c|}{40.32}  & 41.58          \\
\multicolumn{1}{c|}{SASH}                                                      & \multicolumn{1}{c|}{9.40}          & \multicolumn{1}{c|}{14.15}          & \multicolumn{1}{c|}{13.87}          & \multicolumn{1}{c|}{14.96}           & \multicolumn{1}{c|}{7.44}          & \multicolumn{1}{c|}{12.78}          & \multicolumn{1}{c|}{14.26}          & \multicolumn{1}{c|}{15.35}          & \multicolumn{1}{c|}{25.60}  & \multicolumn{1}{c|}{34.21}  & \multicolumn{1}{c|}{38.96}  & 39.74          \\ \hline
\multicolumn{1}{c|}{RAZH-T}                                           & \multicolumn{1}{c|}{\textbf{43.80}} & \multicolumn{1}{c|}{\textbf{45.45}} & \multicolumn{1}{c|}{\textbf{46.26}} & \multicolumn{1}{c|}{\textbf{48.26}} & \multicolumn{1}{c|}{\textbf{12.10}} & \multicolumn{1}{c|}{\textbf{15.87}} & \multicolumn{1}{c|}{\textbf{13.95}} & \multicolumn{1}{c|}{\textbf{15.95}} & \multicolumn{1}{c|}{\textbf{35.27}}              & \multicolumn{1}{c|}{\textbf{40.86}}              & \multicolumn{1}{c|}{\textbf{42.96}}              & \textbf{46.06} \\
\multicolumn{1}{c|}{RAZH}                                            & \multicolumn{1}{c|}{\textbf{43.80}} & \multicolumn{1}{c|}{\textbf{45.45}} & \multicolumn{1}{c|}{\textbf{46.26}} & \multicolumn{1}{c|}{\textbf{48.26}} & \multicolumn{1}{c|}{\textbf{13.13}} & \multicolumn{1}{c|}{\textbf{16.07}} & \multicolumn{1}{c|}{\textbf{16.95}} & \multicolumn{1}{c|}{\textbf{19.95}} & \multicolumn{1}{c|}{\textbf{38.12}}              & \multicolumn{1}{c|}{\textbf{40.93}}              & \multicolumn{1}{c|}{\textbf{44.96}}              & \textbf{46.86} \\ \hline
                                                      &                                      &                                      &                                      &                                      &                                      &                                      &                                      &                                      &                                      &                                      &                                      &                 \\ \hline
\multicolumn{1}{c|}{{DSH}}    & \multicolumn{1}{c|}{{11.76}} & \multicolumn{1}{c|}{{11.07}} & \multicolumn{1}{c|}{{11.46}} & \multicolumn{1}{c|}{{12.30}} & \multicolumn{1}{c|}{{0.71}} & \multicolumn{1}{c|}{{0.88}} & \multicolumn{1}{c|}{{0.89}} & \multicolumn{1}{c|}{{1.07}} & \multicolumn{1}{c|}{{4.10}} & \multicolumn{1}{c|}{{4.23}} & \multicolumn{1}{c|}{{4.93}} & {5.47}       \\
\multicolumn{1}{c|}{DPSH}            & \multicolumn{1}{c|}{16.54}          & \multicolumn{1}{c|}{16.54}          & \multicolumn{1}{c|}{16.94}          & \multicolumn{1}{c|}{16.41}          & \multicolumn{1}{c|}{0.81}          & \multicolumn{1}{c|}{0.81}          & \multicolumn{1}{c|}{0.81}          & \multicolumn{1}{c|}{0.62}          & \multicolumn{1}{c|}{6.96}           & \multicolumn{1}{c|}{8.08}          & \multicolumn{1}{c|}{8.91}          & 10.28                \\
\multicolumn{1}{c|}{DSDH}            & \multicolumn{1}{c|}{14.09}          & \multicolumn{1}{c|}{13.79}          & \multicolumn{1}{c|}{13.76}          & \multicolumn{1}{c|}{13.93}          & \multicolumn{1}{c|}{1.11}          & \multicolumn{1}{c|}{1.54}          & \multicolumn{1}{c|}{1.73}          & \multicolumn{1}{c|}{1.89}          & \multicolumn{1}{c|}{6.74}          & \multicolumn{1}{c|}{8.04}          & \multicolumn{1}{c|}{9.09}          & 9.94                \\
\multicolumn{1}{c|}{DCH}             & \multicolumn{1}{c|}{18.14}          & \multicolumn{1}{c|}{15.04}           & \multicolumn{1}{c|}{17.50}          & \multicolumn{1}{c|}{17.09}          & \multicolumn{1}{c|}{2.64}          & \multicolumn{1}{c|}{3.22}          & \multicolumn{1}{c|}{3.65}          & \multicolumn{1}{c|}{3.21}          & \multicolumn{1}{c|}{6.80}          & \multicolumn{1}{c|}{6.19}          & \multicolumn{1}{c|}{6.26}          & 5.76                \\
\multicolumn{1}{c|}{DDSH}            & \multicolumn{1}{c|}{13.37}          & \multicolumn{1}{c|}{4.52}          & \multicolumn{1}{c|}{15.13}          & \multicolumn{1}{c|}{14.99}           & \multicolumn{1}{c|}{0.55}          & \multicolumn{1}{c|}{0.43}          & \multicolumn{1}{c|}{0.64}          & \multicolumn{1}{c|}{0.40}          & \multicolumn{1}{c|}{6.74}          & \multicolumn{1}{c|}{8.04}          & \multicolumn{1}{c|}{9.09}          & 9.94                \\
\multicolumn{1}{c|}{ADSH}            & \multicolumn{1}{c|}{18.38}          & \multicolumn{1}{c|}{18.97}           & \multicolumn{1}{c|}{18.54}          & \multicolumn{1}{c|}{18.62}          & \multicolumn{1}{c|}{6.04}          & \multicolumn{1}{c|}{8.05}          & \multicolumn{1}{c|}{8.53}          & \multicolumn{1}{c|}{10.19}          & \multicolumn{1}{c|}{17.28}          & \multicolumn{1}{c|}{22.31}          & \multicolumn{1}{c|}{23.31}          & 22.40                \\
\multicolumn{1}{c|}{AMVH}            & \multicolumn{1}{c|}{19.02}          & \multicolumn{1}{c|}{19.21}          & \multicolumn{1}{c|}{19.31}          & \multicolumn{1}{c|}{19.53}          & \multicolumn{1}{c|}{7.01}          & \multicolumn{1}{c|}{8.21}          & \multicolumn{1}{c|}{9.13}          & \multicolumn{1}{c|}{10.65}          & \multicolumn{1}{c|}{18.11}          & \multicolumn{1}{c|}{23.05}          & \multicolumn{1}{c|}{23.67}          & 24.12                \\
\multicolumn{1}{c|}{SitNet}          & \multicolumn{1}{c|}{20.47}          & \multicolumn{1}{c|}{20.70}          & \multicolumn{1}{c|}{20.53}          & \multicolumn{1}{c|}{22.07}          & \multicolumn{1}{c|}{8.80}          & \multicolumn{1}{c|}{11.27}          & \multicolumn{1}{c|}{11.41}          & \multicolumn{1}{c|}{11.67}          & \multicolumn{1}{c|}{23.44}          & \multicolumn{1}{c|}{24.06}          & \multicolumn{1}{c|}{25.49}          & 26.50                \\
\multicolumn{1}{c|}{DRMH}            & \multicolumn{1}{c|}{21.30}          & \multicolumn{1}{c|}{22.21}          & \multicolumn{1}{c|}{23.27}          & \multicolumn{1}{c|}{23.47}          & \multicolumn{1}{c|}{6.79}          & \multicolumn{1}{c|}{8.97}          & \multicolumn{1}{c|}{9.51}          & \multicolumn{1}{c|}{12.49}          & \multicolumn{1}{c|}{27.74}          & \multicolumn{1}{c|}{31.25}          & \multicolumn{1}{c|}{32.39}          & 34.07                \\ \hline
\multicolumn{1}{c|}{RAZH-T}                                           & \multicolumn{1}{c|}{\textbf{43.80}} & \multicolumn{1}{c|}{\textbf{45.45}} & \multicolumn{1}{c|}{\textbf{46.26}} & \multicolumn{1}{c|}{\textbf{48.26}} & \multicolumn{1}{c|}{\textbf{12.10}} & \multicolumn{1}{c|}{\textbf{15.87}} & \multicolumn{1}{c|}{\textbf{13.95}} & \multicolumn{1}{c|}{\textbf{15.95}} & \multicolumn{1}{c|}{\textbf{35.27}}              & \multicolumn{1}{c|}{\textbf{40.86}}              & \multicolumn{1}{c|}{\textbf{42.96}}              & \textbf{46.06} \\
\multicolumn{1}{c|}{RAZH}                                             & \multicolumn{1}{c|}{\textbf{43.80}} & \multicolumn{1}{c|}{\textbf{45.45}} & \multicolumn{1}{c|}{\textbf{46.26}} & \multicolumn{1}{c|}{\textbf{48.26}} & \multicolumn{1}{c|}{\textbf{13.13}} & \multicolumn{1}{c|}{\textbf{16.07}} & \multicolumn{1}{c|}{\textbf{16.95}} & \multicolumn{1}{c|}{\textbf{19.95}} & \multicolumn{1}{c|}{\textbf{38.12}}              & \multicolumn{1}{c|}{\textbf{40.93}}              & \multicolumn{1}{c|}{\textbf{44.96}}              & \textbf{46.86} 

\\ \hline

\end{tabular}
}
\vspace{-0.2cm}
\end{table*}

\textbf{CUB} has 11,788 images, including 200 bird classes and 312 attribute labels. As a zero-shot hashing dataset, it consists of 150 seen classes and 50 unseen classes. Under the experimental settings, the training set contains 4,500 images, with each class consisting of a set of 30 randomly chosen images; the testing set contains 1,500 images, with each class consisting of a set of 30 randomly chosen images; and the remaining 5,788 images are included in the retrieval set.

\textbf{AWA2} has 37,322 images, including 50 animal classes and 85 attribute labels. As a zero-shot hashing dataset, it consists of 40 seen classes and 10 unseen classes. Under the experimental settings, the training set contains 5,000 images, with each class consisting of a set of 100 randomly chosen images; the testing set contains 1,000 images, with each class consisting of a set of 100 randomly chosen images; and the remaining 32,322 images are included in the retrieval set.
\\ \hspace*{\fill} \\
\noindent \emph{B. Metrics}

To evaluate the performance of RAZH, we employ four conventional metrics, specifically, the mean average precision (mAP), precision‒recall curve (PR curve), precision@N curve (P@N curve), and recall@N curve (R@N curve).
\\ \hspace*{\fill} \\
\noindent \emph{C. Experimental Settings}

To make a fair comparison, all unseen-class datasets are randomly selected. For various datasets, the weight coefficients involved in each loss are adjusted accordingly, and the values of $\alpha$ and $\beta$ are determined through a grid search process.

For the three zero-shot hashing benchmark datasets, we specify $d_a=300$ and $d_v=768$. The training procedure employs the Adam gradient descent algorithm, which uses utilizing initial parameters of $\theta_1 = 0.9$, and $\theta_2 = 0.999$, for updating the network parameters. We implement the algorithm via the Python-based PyTorch framework and conduct experiments on an NVIDIA Tesla A100 GPU.
\\ \hspace*{\fill} \\
\noindent \emph{D. Benchmark Dataset Results}

\emph{1) Experimental Results}: To prove the effectiveness of RAZH, we compared it with several zero-shot hashing methods, including the  LSH\cite{andoni2008near}, SH\cite{weiss2008spectral}, ITQ\cite{gong2012iterative}, SDH\cite{shen2015supervised}, FSSH\cite{luo2018fast}, SCDH\cite{chen2020strongly}, TSK\cite{yang2016zero}, AH\cite{xu2017attribute}, OPZH\cite{zhang2019zero}, ASZH\cite{shi2022zero} and SASH\cite{shi2022supervised} linear hashing methods and the DSH\cite{liu2016deep}, DPSH\cite{li2015feature}, DSDH\cite{li2017deep}, DCH\cite{cao2018deep}, DDSH\cite{jiang2018deep}, ADSH\cite{jiang2018asymmetric}, AWVH\cite{lu2020adversarial}, SitNet\cite{guo2017sitnet}, and DRMH\cite{peng2023multi} deep hashing methods.

Table II presents the experimental results obtained on three different datasets: CIFAR10, CUB, and AWA2. The upper part of Table II shows the comparison between the performance-related results of RAZH with those of the linear hashing methods, and the lower part shows the results of the experimental comparison with the existing zero-shot hashing methods. RAZH significantly outperforms the other methods on these three datasets. Compared with the best-performing linear hashing method, AZSH, RAZH improves the mAP averages achieved on the three datasets by 31.49\%, 4.08\%, and 6.22\%, respectively. Among the deep hashing methods, DRMH performs best on the CIFAR10 and AWA2 datasets. Compared with those of the DRMH method, the average mAP of RAZH is improved by 23.38\% and 11.35\%, respectively. On the CUB dataset, SitNet achieves the best results, and RAZH achieves a 5.74\% improvement over SitNet. The experiments demonstrate that the RAZH method has significant advantages over the previously developed zero-shot hashing methods. However, considering the absence of attribute information in many cases, such as those in the CIFAR10 dataset, we employ labels instead of attributes for conducting the experiments (denoted as RAZH-T). On the basis of the results presented in Table II, although RAZH-T slightly lags behind RAZH, it still outperforms the prior methods, which validates the strong generalization ability of the RAZH algorithm.

The mAP produced for hash codes of varying lengths on the AWA2, CIFAR10, and CUB datasets are shown in Fig. 3. The observations indicate a gradual increase in the mAP as the sizes of the hash codes increase.

\begin{figure}[htbp]
	\vspace{-1.5em}
	\centering
	\includegraphics[width=3.4in]{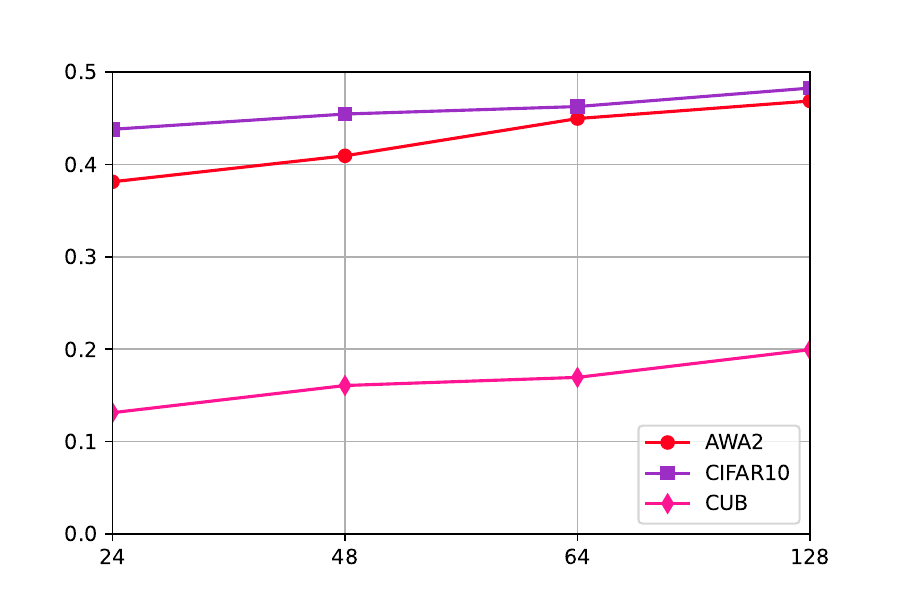}%
	\centering
	\captionsetup{justification=raggedright, singlelinecheck=false}
	\caption{In the three zero-shot hashing datasets, the mAP tends to increase as the hash code length increases. }
	
	\label{fig_sim}
	\vspace{-0.4cm}
\end{figure}

In Fig. 4, we plot the three curves produced on the AWA2 dataset with a 64-bit hash code. RAZH outperforms the vast majority of the other methods in terms of the PR, P@N, and R@N metrics.

\begin{figure*}[htbp]    % 常规操作\begin{figure}开头说明插入图片
		% 后面跟着的[htbp]是图片在文档中放置的位置，也称为浮动体的位置，关于这个我们后面的文章会聊聊，现在不管，照写就是了
		\centering            % 前面说过，图片放置在中间
		\subfloat[PR Curve]   % 第一张子图的下标（注意：注释要写在[]中括号内）
		{
			\label{fig:subfig1}\includegraphics[width=0.30\textwidth]{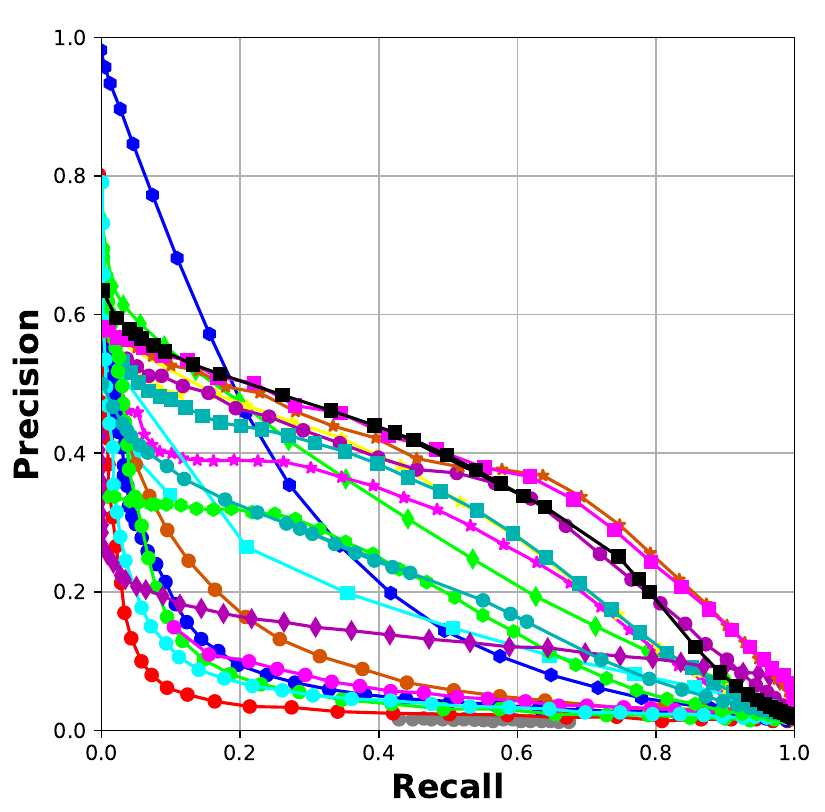}
		}
		\subfloat[P@N Curve]
		{
			\label{fig:subfig2}\includegraphics[width=0.30\textwidth]{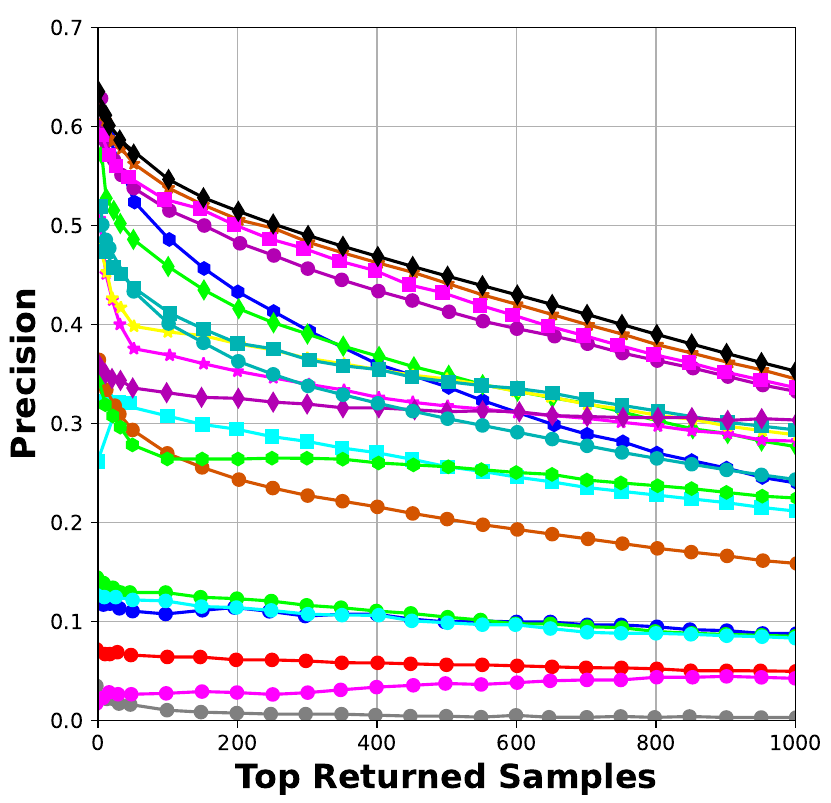}
		}
		\subfloat[R@N Curve]
		{
			\label{fig:subfig3}\includegraphics[width=0.35\textwidth]{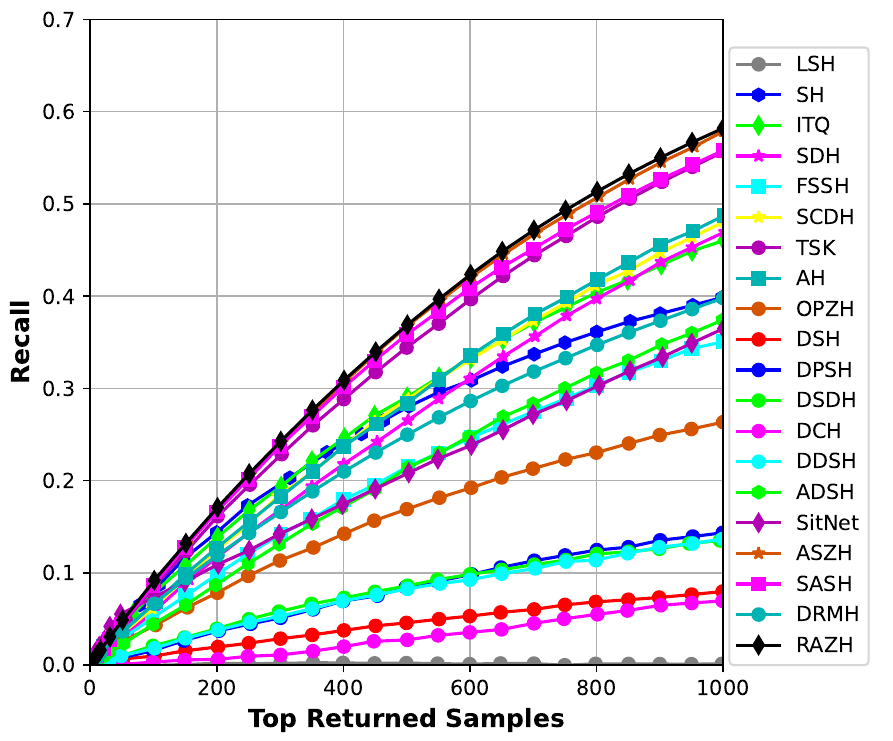}
		}
		\captionsetup{justification=raggedright, singlelinecheck=false}
		\caption{Performance (PR Curve, P@N Curve and R@N Curve) at 64 bits hash codes on the AWA2 dataset.}    % 整个图片的说明，注释写在{}内
		\label{fig:subfig_1}            % 整个图片的标签编号，注意这里跟子图是一样的道理，标签不能重复 
	\end{figure*}

In addition, to illustrate the ability of keeping similarity from original data to hash codes between unseen classes and other classes, Fig. 5 show the confusion matrices for AWA2, CUB, and CIFAR10. Taking AWA2 as an example, when retrieving the unseen class “persian cat”, the closest class is the same class, and the next-closest class is the seen “siamese cat” class. This result indicates that the supervisory information of the “siamese cat” class has been effectively transferred to the “persian cat” class.

\begin{figure*}[htbp]
	\centering
	\includegraphics[width=5.2in]{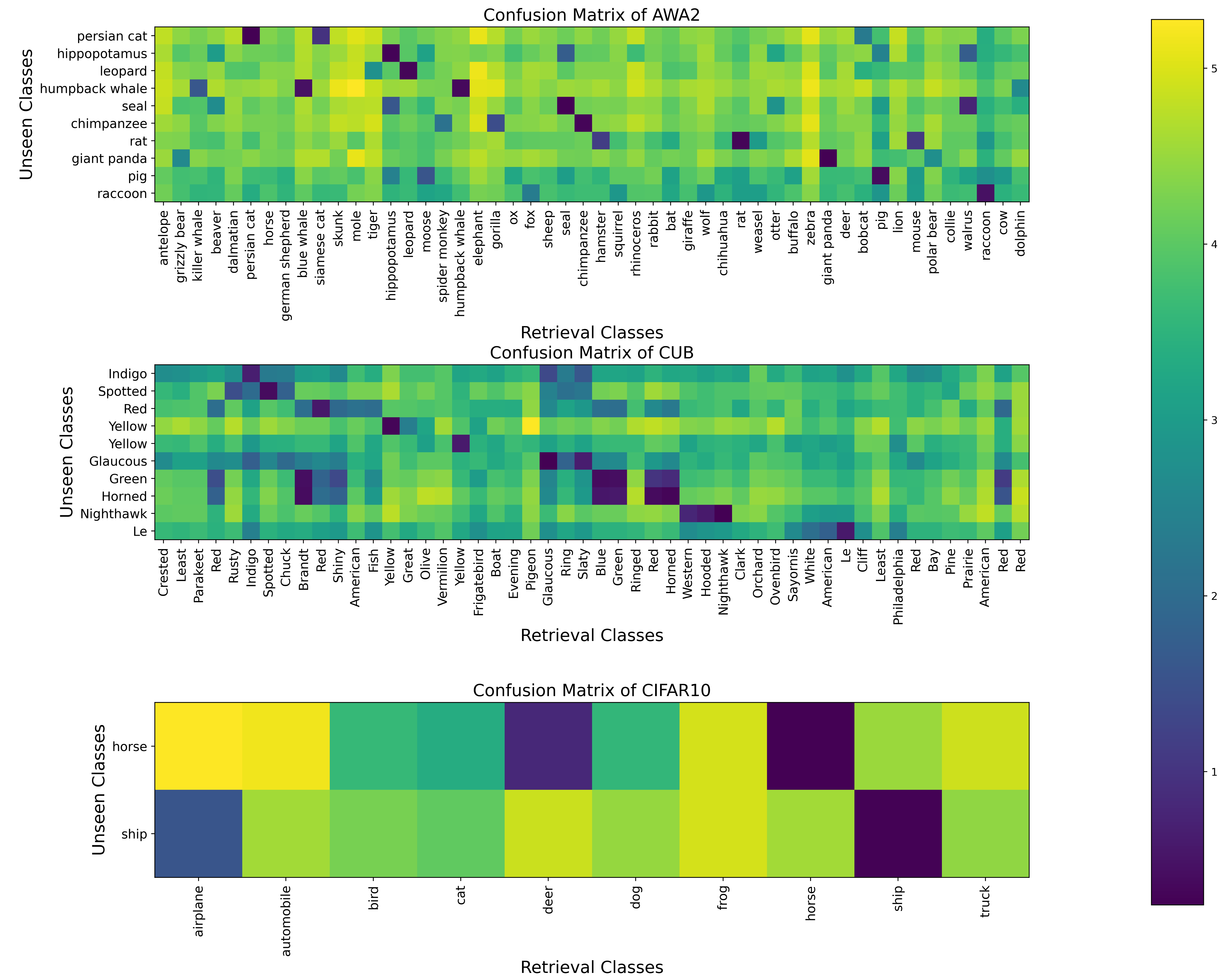}%t
	\centering
	\captionsetup{justification=raggedright, singlelinecheck=false}
	\caption{In the three zero-shot hashing datasets, the confusion matrix of hash code distances between the seen classes and all classes. }
	\label{fig_sim}
\end{figure*}

\begin{table}[]
	\centering
	\caption{}{MODEL PARAMETER COUNT AND FLOPs OF RAZH.}
	\\ [0.2cm]
	\setlength{\tabcolsep}{6.0mm}{
		\renewcommand{\arraystretch}{1.5}
		\begin{tabular}{c|c|c|c}
			\hline
			\textbf{Method} & \textbf{mAP} & \textbf{Params}& \textbf{FLOPs} \\ \hline
			DRMH   & 32.39  & \textbf{60.70M}     & 181.73G    \\ \hline
			RAZH   &  \textbf{44.96} & 85.70M     & \textbf{16.86G}    \\ \hline
	\end{tabular}}

\vspace{-0.8cm}
\end{table}

\begin{table*}[htbp]
	\centering
	\caption{}{RESULTS OF ABLATION STUDY AT DIFFERENT LOSS ON THREE ZERO-SHOT DATASETS.}
	\\ [0.2cm]
	\setlength{\tabcolsep}{5.5mm}{
		\renewcommand{\arraystretch}{1.5}
		\begin{tabular}{c|cc|cc|cc}
			\hline
			\multirow{2}{*}{\textbf{Method}} & \multicolumn{2}{c|}{\textbf{AWA2}}                     & \multicolumn{2}{c|}{\textbf{CUB}}                      & \multicolumn{2}{c}{\textbf{CIFAR10}}                   \\ \cline{2-7} 
			& \multicolumn{1}{c|}{24bits}          & 64bits          & \multicolumn{1}{c|}{24bits}          & 64bits          & \multicolumn{1}{c|}{24bits}          & 64bits          \\ \hline
			$\mathcal{L}_{cls}$                                & \multicolumn{1}{c|}{8.98}          & 8.98         & \multicolumn{1}{c|}{3.72}          & 4.87         & \multicolumn{1}{c|}{25.49}          & 27.24          \\
			$\mathcal{L}_{h}$                                & \multicolumn{1}{c|}{33.98}          & 44.71        & \multicolumn{1}{c|}{10.15}          & 13.94         & \multicolumn{1}{c|}{38.93}          & 42.80        \\
			$\mathcal{L}_{cls}+\alpha\mathcal{L}_h$                               & \multicolumn{1}{c|}{34.55}          & 44.66         & \multicolumn{1}{c|}{13.10}          & 16.11          & \multicolumn{1}{c|}{42.01}          & 44.02          \\
			$\mathcal{L}_{cls}+\alpha\mathcal{L}_h+\beta\mathcal{L}_{rec}$                                & \multicolumn{1}{c|}{\textbf{38.12}} & \textbf{44.96} & \multicolumn{1}{c|}{\textbf{13.13}} & \textbf{16.95} & \multicolumn{1}{c|}{\textbf{43.80}} & \textbf{46.26} \\ \hline
	\end{tabular} }
\end{table*}

% \usepackage{multirow}
\begin{table}[htbp]
	\centering
	\caption{}{mAP(\%) RESULTS WITH DIFFERENT SELECTION RATIO.}
	\\ [0.2cm]
	\setlength{\tabcolsep}{2.0mm}{
		\renewcommand{\arraystretch}{1.5}
		\begin{tabular}{c|cc|cc|cc}
			\hline
			\multirow{2}{*}{\textbf{Mask Ratio}} & \multicolumn{2}{c|}{\textbf{AWA2}}   & \multicolumn{2}{c|}{\textbf{CUB}}    & \multicolumn{2}{c}{\textbf{CIFAR10}} \\ \cline{2-7} 
			& \multicolumn{1}{c|}{24bits} & 64bits & \multicolumn{1}{c|}{24bits} & 64bits & \multicolumn{1}{c|}{24bits} & 64bits \\ \hline
			0.25                                 & \multicolumn{1}{c|}{37.13} & 44.36 & \multicolumn{1}{c|}{12.67} & \textbf{18.29} & \multicolumn{1}{c|}{43.20} & \textbf{46.80} \\
			0.50                                  & \multicolumn{1}{c|}{\textbf{38.12}} & 44.96 & \multicolumn{1}{c|}{\textbf{13.13}} & 16.95 & \multicolumn{1}{c|}{\textbf{43.80}} & 46.26 \\
			0.75                                 & \multicolumn{1}{c|}{36.33} & \textbf{45.19} & \multicolumn{1}{c|}{11.66} & 16.17 & \multicolumn{1}{c|}{42.99} & 46.23 \\ \hline
	\end{tabular} }
\end{table}

\begin{table}[htbp]
	\centering
	\caption{}{mAP(\%) RESULTS USING DIFFERENT COMBINATIONS  OF MODULES AT AWA2, CUB AND CIFAR10.}
	\\ [0.2cm]
	\setlength{\tabcolsep}{1.3mm}{
		\renewcommand{\arraystretch}{1.5}
		\begin{tabular}{c|l|llllll}
			\hline
			\multirow{2}{*}{\textbf{Method}} & \multicolumn{1}{c|}{\multirow{2}{*}{\textbf{Backbone}}} & \multicolumn{2}{c|}{\textbf{AWA2}}                                & \multicolumn{2}{c|}{\textbf{CUB}}                                 & \multicolumn{2}{c}{\textbf{CIFAR10}}                             \\ \cline{3-8} 
			& \multicolumn{1}{c|}{}                          & \multicolumn{1}{c|}{24bits} & \multicolumn{1}{c|}{64bits} & \multicolumn{1}{c|}{24bits} & \multicolumn{1}{c|}{64bits} & \multicolumn{1}{c|}{24bits} & \multicolumn{1}{c}{64bits} \\ \hline
			RAZH-C                   & \multirow{4}{*}{Transformer}                  &  \multicolumn{1}{c|}{35.02}                      &  \multicolumn{1}{c|}{43.09}                      &  \multicolumn{1}{c|}{11.15}                      &  \multicolumn{1}{c|}{15.90}                     &  \multicolumn{1}{c|}{40.46}                   &  \multicolumn{1}{c}{46.23}                     \\
			RAZH-A                   &                  &  \multicolumn{1}{c|}{36.70}                    &  \multicolumn{1}{c|}{44.30}                     &  \multicolumn{1}{c|}{10.05}                      &  \multicolumn{1}{c|}{13.90}                     &  \multicolumn{1}{c|}{30.53}                     &  \multicolumn{1}{c}{32.75}                    \\
			RAZH-D                   &                                               &  \multicolumn{1}{c|}{35.87}                      &  \multicolumn{1}{c|}{44.77}                  &  \multicolumn{1}{c|}{11.02}                    &  \multicolumn{1}{c|}{15.57}                     &  \multicolumn{1}{c|}{40.50}                     &  \multicolumn{1}{c}{44.75}                     \\
			RAZH                     &                                               &  \multicolumn{1}{c|}{\textbf{38.12}}                      &  \multicolumn{1}{c|}{\textbf{44.96}}                  &  \multicolumn{1}{c|}{\textbf{13.13}}                      &  \multicolumn{1}{c|}{\textbf{16.95}}                    &  \multicolumn{1}{c|}{\textbf{43.80}}                     &  \multicolumn{1}{c}{\textbf{46.26}}                     \\ \hline
	\end{tabular} }
	\vspace{-0.4cm}
\end{table}

% \usepackage{multirow}
\begin{table}[h]
	\centering
	\caption{}{mAP(\%) RESULTS OF COMAE and RAZH-F.}
	\\ [0.2cm]
	\setlength{\tabcolsep}{1.2mm}{
		\renewcommand{\arraystretch}{1.5}
		\begin{tabular}{c|clll|cclc}
			\hline
			\multirow{2}{*}{\textbf{Methods}} & \multicolumn{4}{c|}{\textbf{AWA2}}         & \multicolumn{4}{c}{\textbf{CUB}}                                                   \\ \cline{2-9} 
			&  \multicolumn{1}{c|}{24bits} &  \multicolumn{1}{c|}{48bits} &  \multicolumn{1}{c|}{64bits} &  \multicolumn{1}{c|}{128bits} &  \multicolumn{1}{c|}{24bits} & \multicolumn{1}{c|}{48bits} &  \multicolumn{1}{c|}{64bits} & \multicolumn{1}{c}{128bits} \\ \hline
			COMAE                    &  \multicolumn{1}{c|}{38.19}  &  \multicolumn{1}{c|}{47.92}  &  \multicolumn{1}{c|}{51.33}  &  \multicolumn{1}{c|}{54.65}  &  \multicolumn{1}{c|}{11.36}  &  \multicolumn{1}{c|}{17.77}                      &  \multicolumn{1}{c|}{19.94}  &  \multicolumn{1}{c}{25.19}                      \\
			RAZH-F                   &  \multicolumn{1}{c|}{\textbf{50.46}}  &  \multicolumn{1}{c|}{\textbf{59.10}}  &  \multicolumn{1}{c|}{\textbf{61.54}}  &  \multicolumn{1}{c|}{\textbf{67.00}}  &  \multicolumn{1}{c|}{\textbf{16.33}}  &  \multicolumn{1}{c|}{\textbf{23.10}}                      &  \multicolumn{1}{c|}{\textbf{25.75}}  &  \multicolumn{1}{c}{\textbf{30.02}}                      \\ \hline
	\end{tabular}}
	\vspace{-0.4cm}
\end{table}

We analyze the parameter counts and FLOPs of the models, as shown in Table III. Although RAZH has a slightly higher parameter count than DRMH does, the latter requires the use of the Faster R-CNN for object detection, which significantly increases its FLOPs. The complexity of the Faster R-CNN results in lower inference efficiency for DRMH. In contrast, RAZH efficiently aligns image patches and attributes without relying on additional object detection models, thereby significantly enhancing its overall performance and runtime efficiency.

	\begin{figure*}[b] 
		\vspace{-0.8cm}
		% 常规操作\begin{figure}开头说明插入图片
			% 后面跟着的[htbp]是图片在文档中放置的位置，也称为浮动体的位置，关于这个我们后面的文章会聊聊，现在不管，照写就是了
			\centering            % 前面说过，图片放置在中间
			\subfloat[24 bits]   % 第一张子图的下标（注意：注释要写在[]中括号内）
			{
				\label{fig:subfig1}\includegraphics[width=0.20\textwidth]{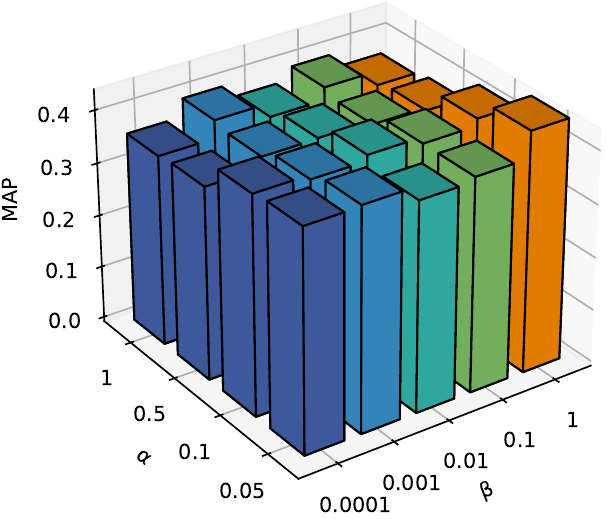}
			}
			\subfloat[48 bits]
			{
				\label{fig:subfig2}\includegraphics[width=0.20\textwidth]{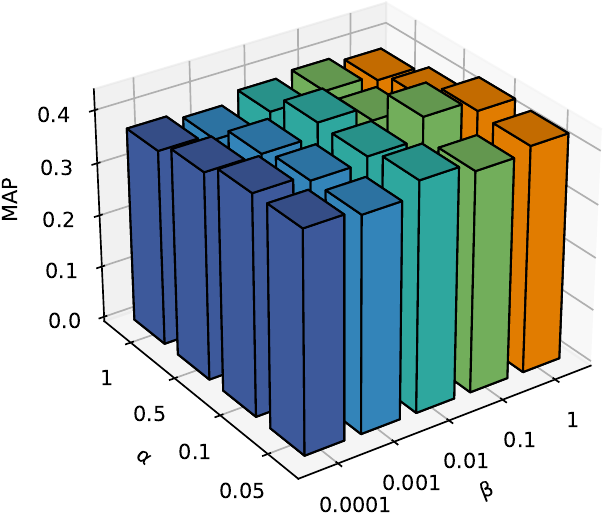}
			}
			\subfloat[64 bits]
			{
				\label{fig:subfig3}\includegraphics[width=0.20\textwidth]{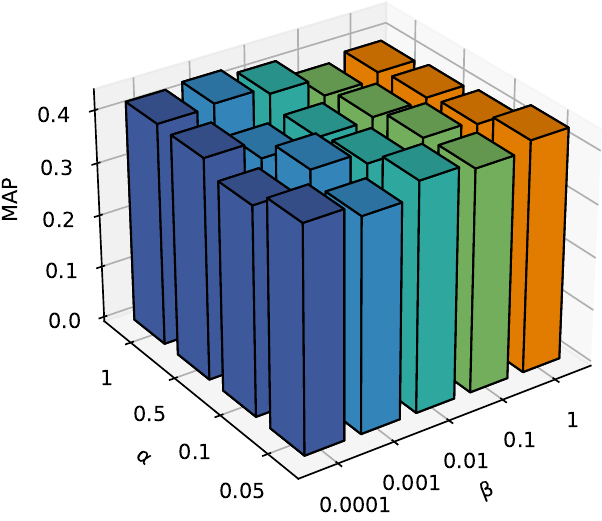}
			}
			\subfloat[128 bits]
			{
				\label{fig:subfig3}\includegraphics[width=0.20\textwidth]{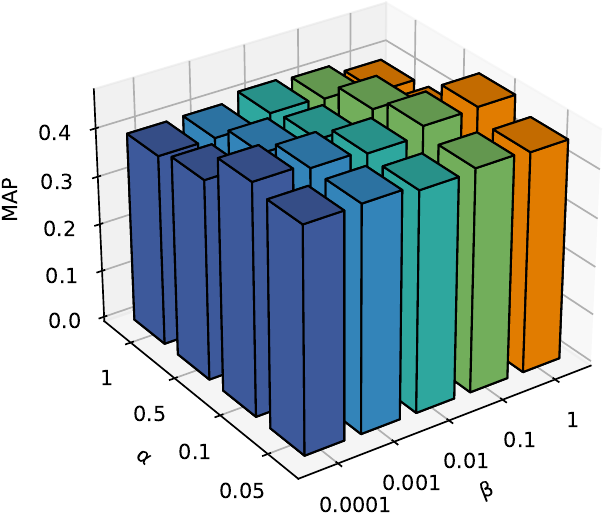}
			}
			\caption{Performance of mAP@5000 under Various Configurations of $\alpha$ and $\beta$ on the CIFAR10.}    % 整个图片的说明，注释写在{}内
			\label{fig:subfig_1}            % 整个图片的标签编号，注意这里跟子图是一样的道理，标签不能重复 
			\vspace{-0.6cm}
		\end{figure*}

		\begin{figure*}[h]   
			\vspace{-0.6cm}
			% 常规操作\begin{figure}开头说明插入图片
				% 后面跟着的[htbp]是图片在文档中放置的位置，也称为浮动体的位置，关于这个我们后面的文章会聊聊，现在不管，照写就是了
				\centering            % 前面说过，图片放置在中间
				\subfloat[24 bits]   % 第一张子图的下标（注意：注释要写在[]中括号内）
				{
					\label{fig:subfig1}\includegraphics[width=0.20\textwidth]{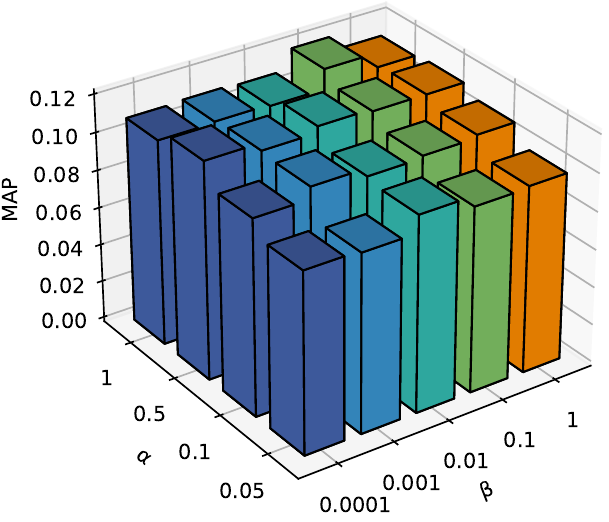}
				}
				\subfloat[48 bits]
				{
					\label{fig:subfig2}\includegraphics[width=0.20\textwidth]{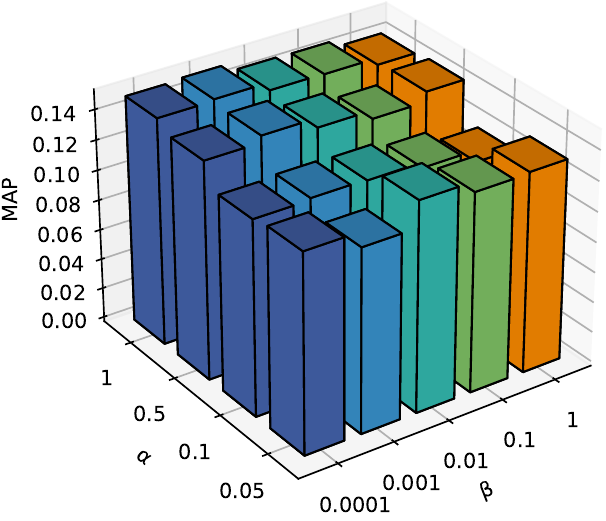}
				}
				\subfloat[64 bits]
				{
					\label{fig:subfig3}\includegraphics[width=0.20\textwidth]{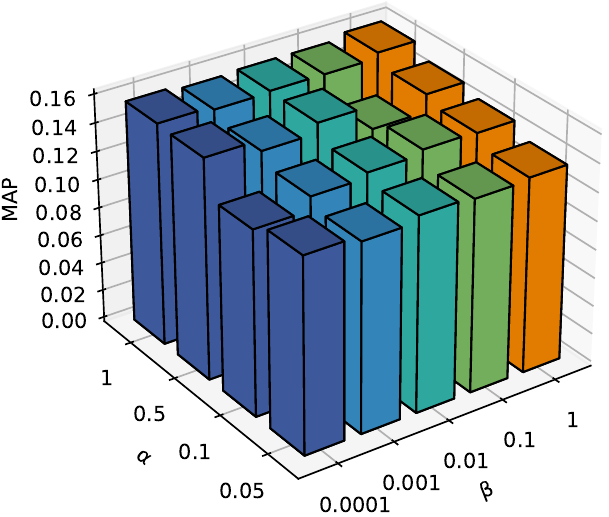}
				}
				\subfloat[128 bits]
				{
					\label{fig:subfig3}\includegraphics[width=0.20\textwidth]{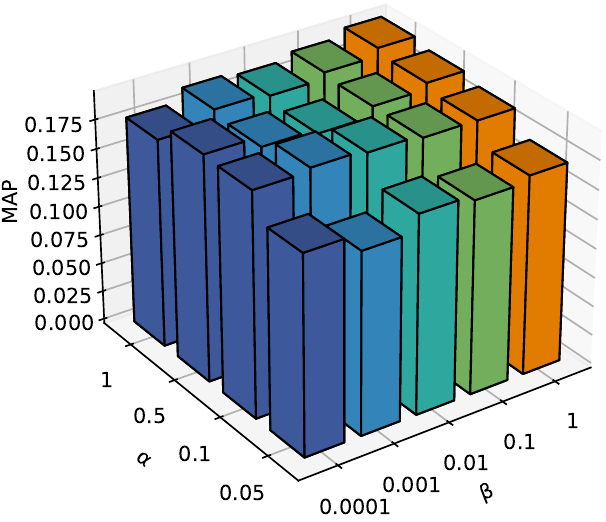}
				}
				\caption{Performance of mAP@5000 under Various Configurations of $\alpha$ and $\beta$ on the CUB.}    % 整个图片的说明，注释写在{}内
				\label{fig:subfig_1}  
				\vspace{-2.0em}          % 整个图片的标签编号，注意这里跟子图是一样的道理，标签不能重复 
			\end{figure*}

		\begin{figure*}[h]    % 常规操作\begin{figure}开头说明插入图片
				% 后面跟着的[htbp]是图片在文档中放置的位置，也称为浮动体的位置，关于这个我们后面的文章会聊聊，现在不管，照写就是了
				\centering            % 前面说过，图片放置在中间
				\subfloat[24 bits]   % 第一张子图的下标（注意：注释要写在[]中括号内）
				{
					\label{fig:subfig1}\includegraphics[width=0.20\textwidth]{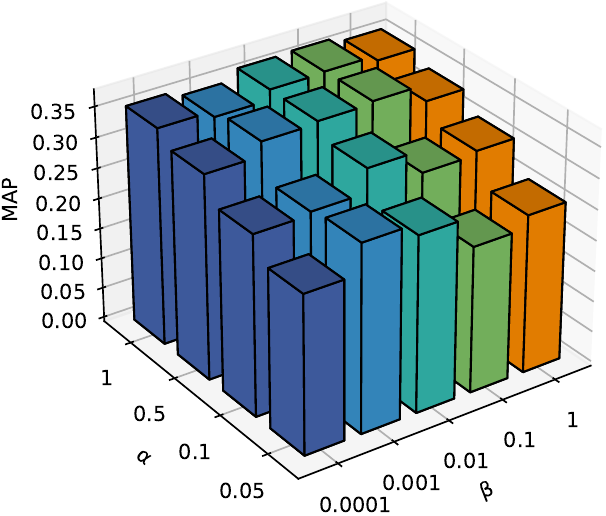}
				}
				\subfloat[48 bits]
				{
					\label{fig:subfig2}\includegraphics[width=0.20\textwidth]{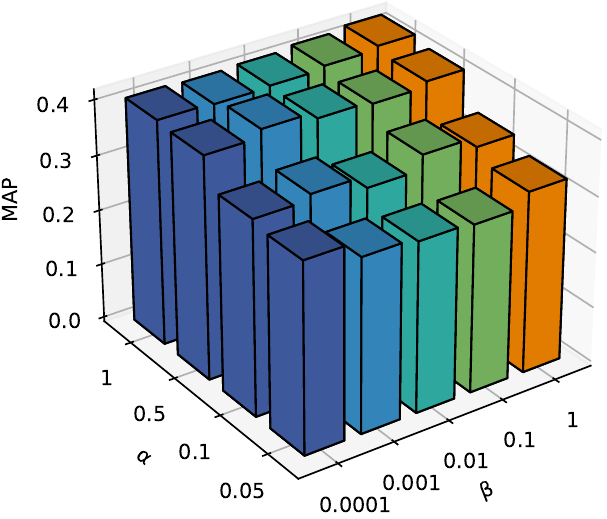}
				}
				\subfloat[64 bits]
				{
					\label{fig:subfig3}\includegraphics[width=0.20\textwidth]{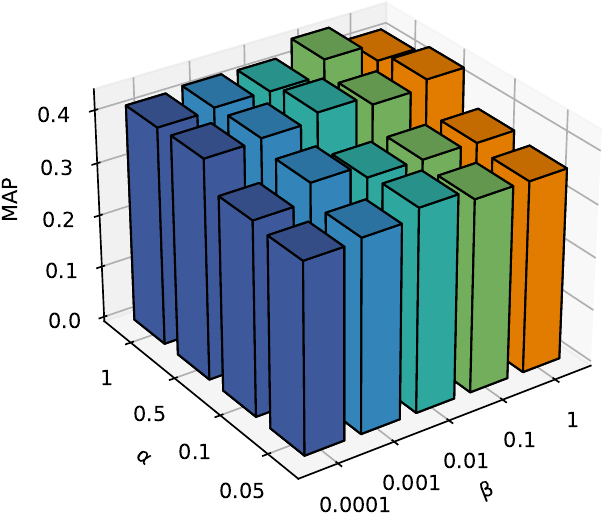}
				}
				\subfloat[128 bits]
				{
					\label{fig:subfig3}\includegraphics[width=0.20\textwidth]{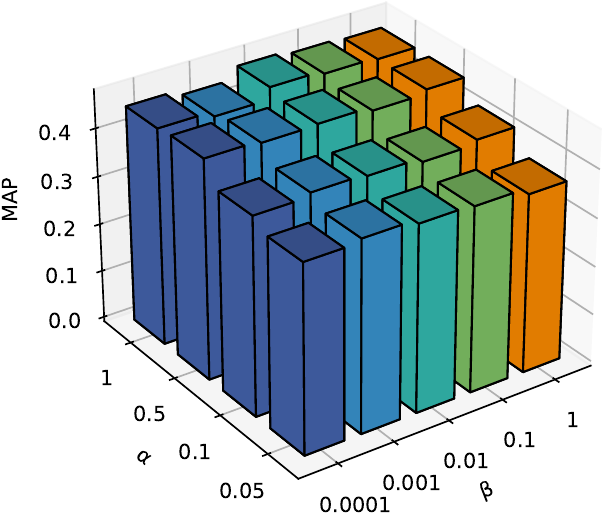}
				}
				\caption{Performance of mAP@5000 under Various Configurations of $\alpha$ and $\beta$ on the AWA2.}    % 整个图片的说明，注释写在{}内
				\label{fig:subfig_1}   
				% 整个图片的标签编号，注意这里跟子图是一样的道理，标签不能重复 
				
			\end{figure*}
			
			\begin{figure*}[h]
				\vspace{-0.6cm}
				    % 常规操作\begin{figure}开头说明插入图片
					% 后面跟着的[htbp]是图片在文档中放置的位置，也称为浮动体的位置，关于这个我们后面的文章会聊聊，现在不管，照写就是了
					\centering    
					\captionsetup{justification=raggedright}        % 前面说过，图片放置在中间
					\subfloat[AWA2]   % 第一张子图的下标（注意：注释要写在[]中括号内）
					{
						\label{fig:subfig1}\includegraphics[width=0.25\textwidth]{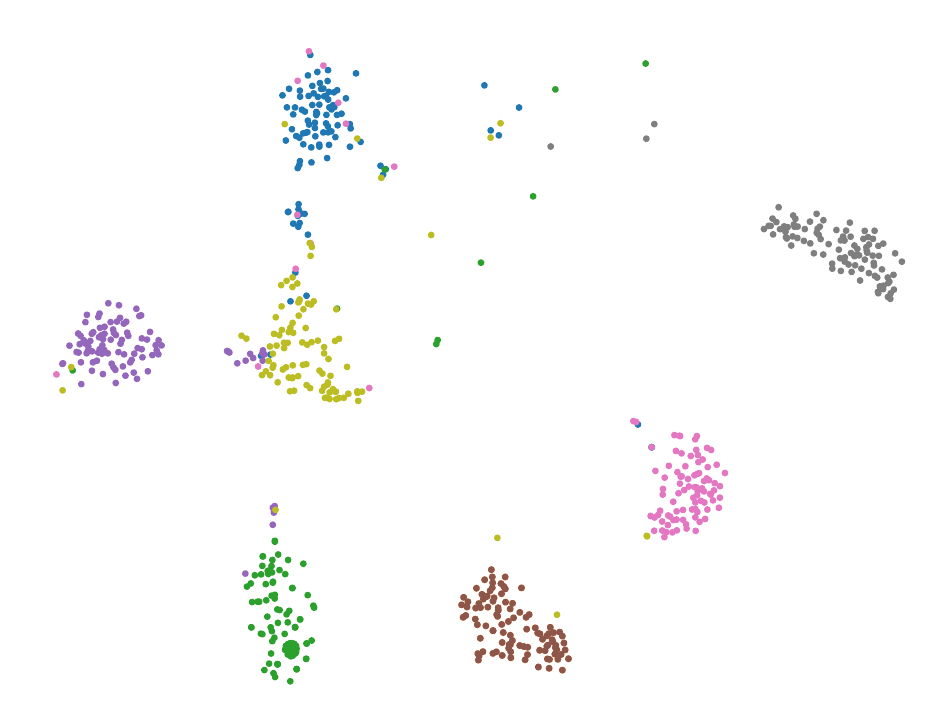}
					} \hspace{0.5cm}
					\subfloat[CIFAR10]
					{
						\label{fig:subfig2}\includegraphics[width=0.25\textwidth]{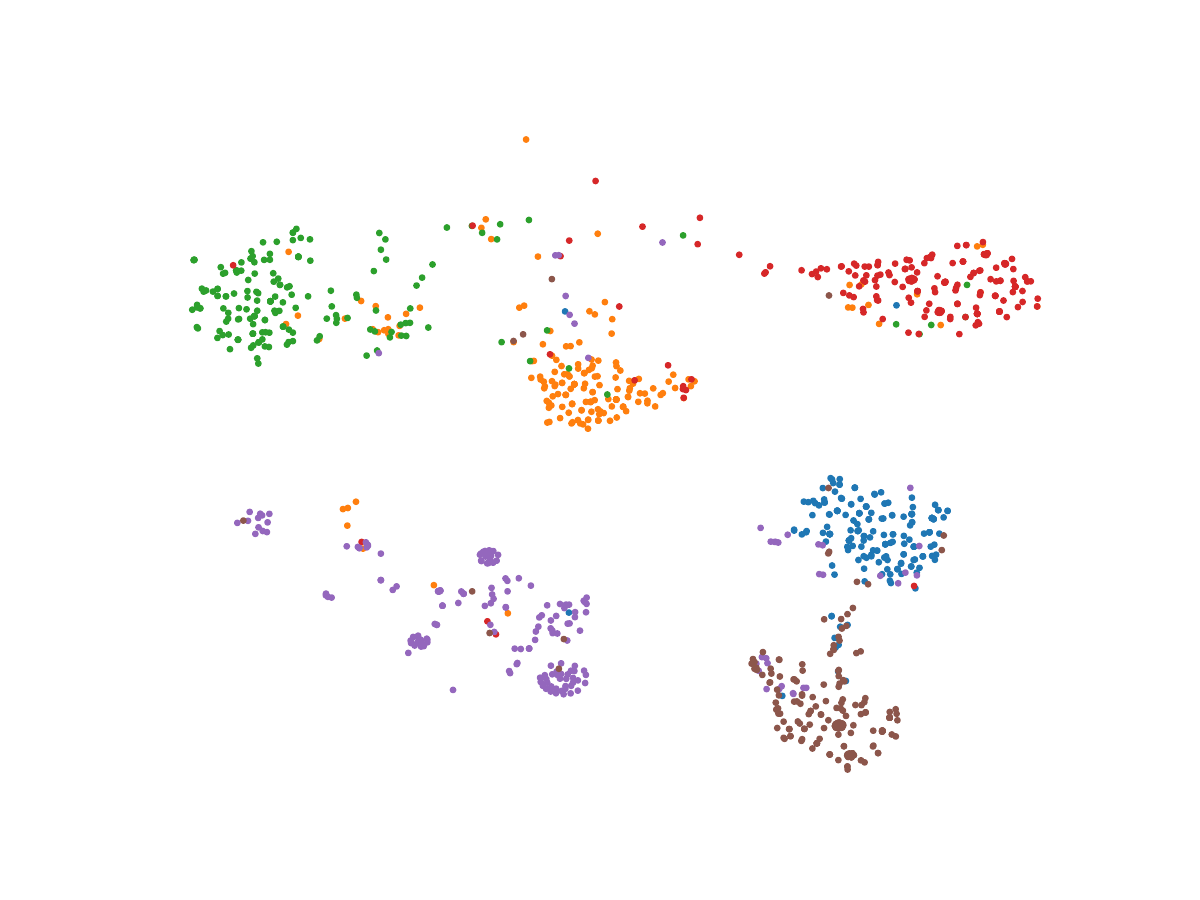}
					}\hspace{0.5cm}
					\subfloat[CUB]
					{
						\label{fig:subfig3}\includegraphics[width=0.25\textwidth]{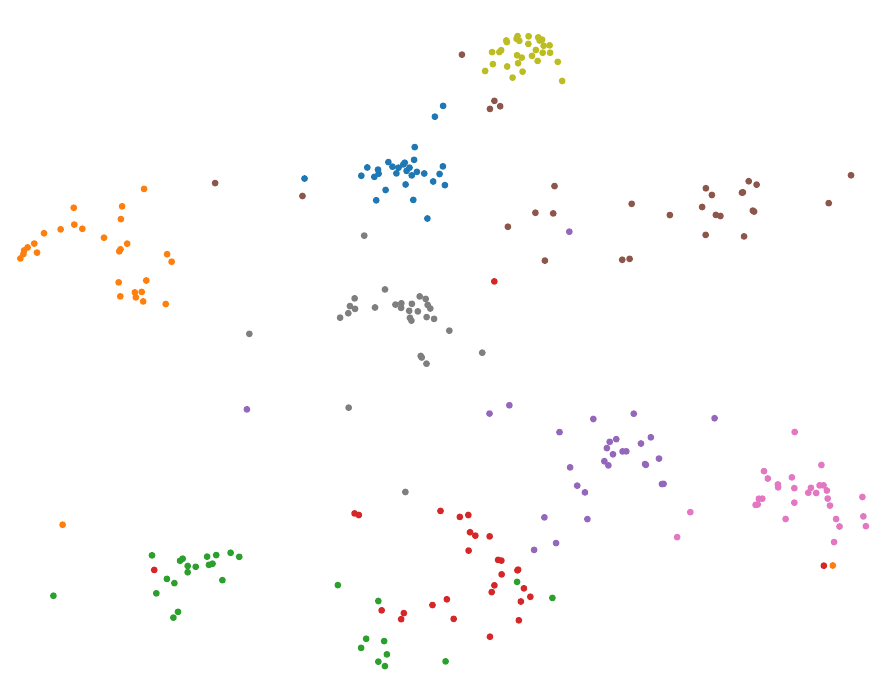}
					}
					\caption{The t-SNE scatterplot shows the distribution of 64-bit hash codes obtained from randomly selected classes images from the AWA2, CIFAR10 and CUB datasets. Different coloured dots represent different classes.}    % 整个图片的说明，注释写在{}内
					\label{fig:subfig_1}            % 整个图片的标签编号，注意这里跟子图是一样的道理，标签不能重复 
				\end{figure*}

\emph{2) Ablation Study}: We conducted loss-based ablation experiments to assess how each element of the loss function affects the obtained retrieval results. We used three losses, namely, $\mathcal{L}_{cls}$, $\mathcal{L}_{h}$ and $\mathcal{L}_{rec}$, by varying the loss composition to observe the differences in mAP differences induced on the three benchmark datasets. As shown in Table IV, the enhanced  results are clearly influenced by the collective contribution of all the loss components. When only $\mathcal{L}_{cls}$ is used for supervision, the results are poor; when only $\alpha\mathcal{L}_{h}$ is used, the results improve significantly; when $\mathcal{L}_{cls}$ and $\alpha\mathcal{L}_{h}$ are jointly supervised, the retrieval results improve further; and when $\beta\mathcal{L}_{rec}$ is added, the retrieval results improve even further. The significance of hash loss can be observed as the most prominent aspect of RAZH.

We perform ablation experiments to analyze the effects of different modules. We first test the effects of different selection rates on the resulting retrieval performance. We use mask ratios of 0.25, 0.50, and 0.75, and the corresponding performances are shown in Table V. When the selection rate is 0.75, less image information is retained; thus, the mAP decreases. When the selection rate is 0.25, more image information is retained, and the corresponding attribute information is less abundant; thus, the model performs poorly on the fine-grained CUB dataset. When the selection rate is 0.50, the image information and attribute information are relatively balanced, and the method can produce better results on the fine-grained dataset. The experimental results demonstrate that RAZH highlights different aspects under various selection rates, thereby demonstrating its adaptability to diverse datasets for the purpose of learning.

To assess the effectiveness of each module, we conduct a series of experiments. First, we remove the clustering algorithm (denoted as RAZH-C) and replace each patch with its closest attribute vector. Next, we eliminate the attribute embedding module, retaining only the selected patch branch (denoted as RAZH-A). Finally, we use two randomly selected dual branches with different patches (denoted as RAZH-D). As shown in Table VI, RAZH slightly outperforms RAZH-C, RAZH-A and RAZH-D, indicating the effectiveness of the clustering algorithm, attribute embedding module, and dual-branch structure.

In terms of performance, COMAE\cite{2024Comae} uses a pretrained ResNet101 as its backbone and incorporates class supervision information. However, there is an ongoing debate about whether this approach fully aligns with the definition of zero-shot hashing (ZSH), as it relies on pretrained class information for supervised learning. To conduct a fair comparison, we replace the backbone with a version (ViT-finetune)\cite{dosovitskiy2020image} that incorporates class information and compare it with COMAE. The results in Table VII show that our method outperforms COMAE.

\begin{figure*}[pbt] 
	\vspace{-0.4cm}
	
	    %[pbt]：表格最优位置
	\centering
	
	\setlength{\extrarowheight}{3pt}{
		\renewcommand{\arraystretch}{1.2}
		\begin{tabular}{ccccc}
			
			\textbf{Input} & \textbf{Cluster}   & & \textbf{Input} & \textbf{Cluster}  \\ 
			\begin{minipage}[b]{0.35\columnwidth}\vspace{2pt}
				\centering
				\raisebox{-.42\height}{\includegraphics[width=0.8\textwidth]{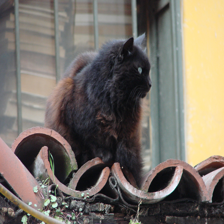}}
			\end{minipage}                    &  
			\begin{minipage}[b]{0.35\columnwidth}
				\centering
				\raisebox{-.42\height}{\includegraphics[width=0.8\textwidth]{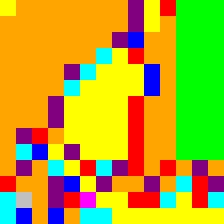}}
			\end{minipage}   &  & \begin{minipage}[b]{0.35\columnwidth}
			\centering
				\raisebox{-.42\height}{\includegraphics[width=0.8\textwidth]{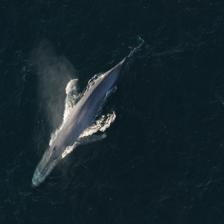}}
			\end{minipage}                    &
			\begin{minipage}[b]{0.35\columnwidth}
				\centering
				\raisebox{-.42\height}{\includegraphics[width=0.8\textwidth]{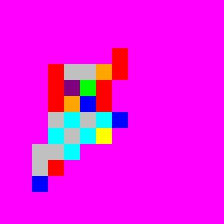}}
			\end{minipage}   \\ 
			
			\begin{minipage}[b]{0.35\columnwidth}\vspace{2pt}
				\centering
				\raisebox{-.42\height}{\includegraphics[width=0.8\textwidth]{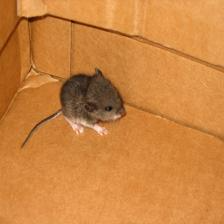}}
			\end{minipage}                    &  
			\begin{minipage}[b]{0.35\columnwidth}
				\centering
				\raisebox{-.42\height}{\includegraphics[width=0.8\textwidth]{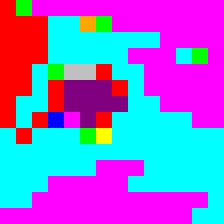}}
			\end{minipage}   &  & \begin{minipage}[b]{0.35\columnwidth}
			\centering
				\raisebox{-.42\height}{\includegraphics[width=0.8\textwidth]{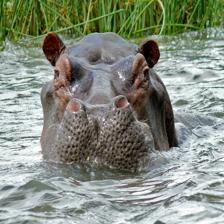}}
			\end{minipage}                    &
			\begin{minipage}[b]{0.35\columnwidth}
				\centering
				\raisebox{-.42\height}{\includegraphics[width=0.8\textwidth]{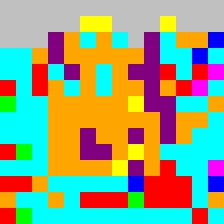}}
			\end{minipage} 
			 \\  
			 \begin{minipage}[b]{0.35\columnwidth}\vspace{2pt}
			 	\centering
			 	\raisebox{-.42\height}{\includegraphics[width=0.8\textwidth]{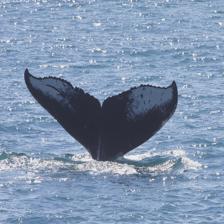}}
			 \end{minipage}                    &  
			 \begin{minipage}[b]{0.35\columnwidth}
			 	\centering
			 	\raisebox{-.42\height}{\includegraphics[width=0.8\textwidth]{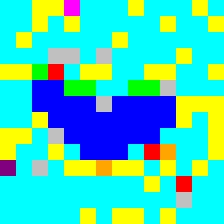}}
			 \end{minipage}   &  & \begin{minipage}[b]{0.35\columnwidth}
			 \centering
			 	\raisebox{-.42\height}{\includegraphics[width=0.8\textwidth]{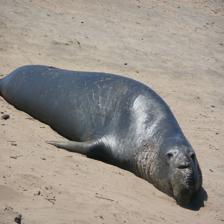}}
			 \end{minipage}                    &
			 \begin{minipage}[b]{0.35\columnwidth}
			 	\centering
			 	\raisebox{-.42\height}{\includegraphics[width=0.8\textwidth]{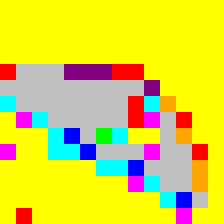}}
			 \end{minipage} \\

	\end{tabular}}
	\caption{Visualization of RAZH cluster. }
\end{figure*}

\begin{figure*}[pbt]     %[pbt]：表格最优位置
\centering
\setlength{\tabcolsep}{2pt}
\setlength{\extrarowheight}{3pt}{
\renewcommand{\arraystretch}{1.5}
\begin{tabular}{ccccccc}

\textbf{Input} & \textbf{Feature}  & \textbf{Prediction} & & \textbf{Input} & \textbf{Feature}  & \textbf{Prediction} \\ 
\begin{minipage}[b]{0.35\columnwidth}\vspace{2pt}\centering
		\raisebox{-.42\height}{\includegraphics[width=0.8\textwidth]{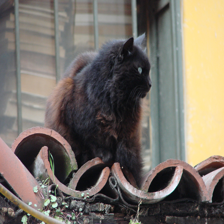}}
\end{minipage}                    &  
\begin{minipage}[b]{0.35\columnwidth}\centering
		\raisebox{-.42\height}{\includegraphics[width=0.8\textwidth]{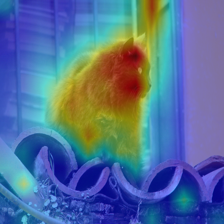}}
\end{minipage}  &  
\makecell[l]{~\includegraphics[width=0.018\textwidth]{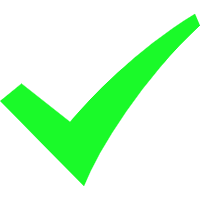}\textbf{black} \\ ~\includegraphics[width=0.018\textwidth]{pics/dui.png}\textbf{brown} \\ ~\includegraphics[width=0.018\textwidth]{pics/dui.png}\textbf{furry} \\ ~\includegraphics[width=0.018\textwidth]{pics/dui.png}\textbf{paws}\\ ~\includegraphics[width=0.018\textwidth]{pics/dui.png}\textbf{pads}} &  & \begin{minipage}[b]{0.35\columnwidth}\centering
		\raisebox{-.42\height}{\includegraphics[width=0.8\textwidth]{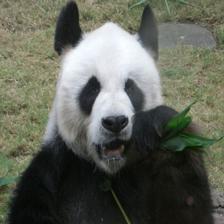}}
\end{minipage}                    &
\begin{minipage}[b]{0.35\columnwidth}\centering
		\raisebox{-.42\height}{\includegraphics[width=0.8\textwidth]{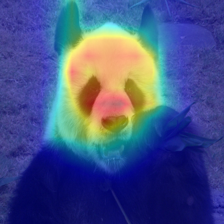}}
\end{minipage}  &
\makecell[l]{~\includegraphics[width=0.018\textwidth]{pics/dui.png}\textbf{black}\\ ~\includegraphics[width=0.018\textwidth]{pics/dui.png}\textbf{white}\\ ~\includegraphics[width=0.018\textwidth]{pics/dui.png}\textbf{furry} \\ ~\includegraphics[width=0.018\textwidth]{pics/dui.png}\textbf{chew teeth} }  \\ 

\begin{minipage}[b]{0.35\columnwidth}\vspace{4pt}\centering
		\raisebox{-.5\height}{\includegraphics[width=0.8\textwidth]{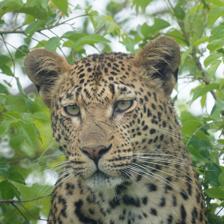}}
\end{minipage}                    &  
\begin{minipage}[b]{0.35\columnwidth}\centering
		\raisebox{-.5\height}{\includegraphics[width=0.8\textwidth]{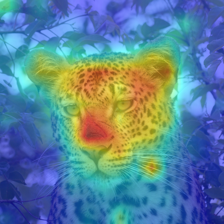}}
\end{minipage}  &
\makecell[l]{~\includegraphics[width=0.018\textwidth]{pics/dui.png}\textbf{black} \\ ~\includegraphics[width=0.018\textwidth]{pics/dui.png}\textbf{spots}\\ ~\includegraphics[width=0.018\textwidth]{pics/dui.png}\textbf{furry}\\ {~\includegraphics[width=0.018\textwidth]{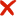}\textbf{stripes}}\\ {~\includegraphics[width=0.018\textwidth]{pics/cuo.png}\textbf{buck teeth}}} & &\begin{minipage}[b]{0.35\columnwidth}\centering
		\raisebox{-.5\height}{\includegraphics[width=0.8\textwidth]{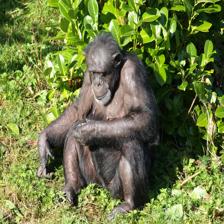}}
\end{minipage}                    &  
\begin{minipage}[b]{0.35\columnwidth}\centering
		\raisebox{-.5\height}{\includegraphics[width=0.8\textwidth]{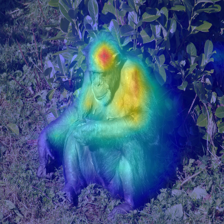}}
\end{minipage}  &
\makecell[l]{~\includegraphics[width=0.018\textwidth]{pics/dui.png}\textbf{gray}\textbf{} \\ ~\includegraphics[width=0.018\textwidth]{pics/dui.png}\textbf{furry}\\ ~\includegraphics[width=0.018\textwidth]{pics/dui.png}\textbf{hands}\\{~\includegraphics[width=0.018\textwidth]{pics/cuo.png}\textbf{stalker}} }  \\  
\begin{minipage}[b]{0.35\columnwidth}\vspace{4pt}\centering
		\raisebox{-.5\height}{\includegraphics[width=0.8\textwidth]{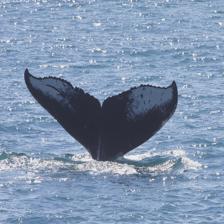}}
\end{minipage}                    &  
\begin{minipage}[b]{0.35\columnwidth}\centering
		\raisebox{-.5\height}{\includegraphics[width=0.8\textwidth]{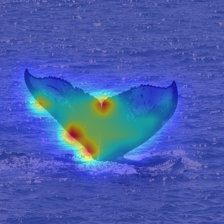}}
\end{minipage}  &
\makecell[l]{~\includegraphics[width=0.018\textwidth]{pics/dui.png}\textbf{flippers} \\~\includegraphics[width=0.018\textwidth]{pics/dui.png}\textbf{water}\\ ~\includegraphics[width=0.018\textwidth]{pics/dui.png}\textbf{blue}\\ {~\includegraphics[width=0.018\textwidth]{pics/cuo.png}\textbf{hooves}} \\{{~\includegraphics[width=0.018\textwidth]{pics/cuo.png}\textbf{pads}}} } & &\begin{minipage}[b]{0.35\columnwidth}\centering
		\raisebox{-.5\height}{\includegraphics[width=0.8\textwidth]{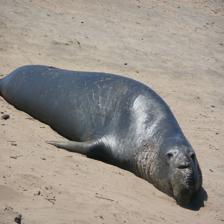}}
\end{minipage}                    &  
\begin{minipage}[b]{0.35\columnwidth}\centering
		\raisebox{-.5\height}{\includegraphics[width=0.8\textwidth]{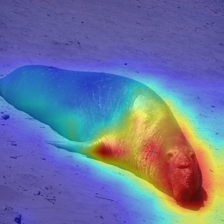}}
\end{minipage}  &
\makecell[l]{~\includegraphics[width=0.018\textwidth]{pics/dui.png}\textbf{gray}\\ ~\includegraphics[width=0.018\textwidth]{pics/dui.png}\textbf{hairless} \\ ~\includegraphics[width=0.018\textwidth]{pics/dui.png}\textbf{flippers}\\{~\includegraphics[width=0.018\textwidth]{pics/cuo.png}\textbf{longleg}} \\{~\includegraphics[width=0.018\textwidth]{pics/cuo.png}\textbf{bipedal}}   }  \\  
\end{tabular}}
\captionsetup{justification=raggedright, singlelinecheck=false}
\caption{Visualization of RAZH attention. Prediction column represents the attributes of the patches mapped to the attribute space, with red being incorrect correspondences and black being correct correspondences.}
\end{figure*}

\emph{3) Parameter Sensitivity Analysis}: Sensitivity experiments are performed on three datasets via fixed classification weights, while varying the hyperparameters $\alpha$ and $\beta$. These hyperparameters denote the weights of the hash and reconstruction losses in the total loss function, respectively. Fig. 6 to Fig. 8 show the mAP results obtained with various hash code lengths across the three benchmark datasets. Finally, we successfully acquired the values of $\alpha$ and $\beta$ corresponding to the optimal retrieval outcomes observed across the various hash code lengths, as depicted in Table VIII. For the AWA2 and CUB datasets, the $\mathcal{L}_{h}$ loss is more important, with the mAP decreasing as alpha decreases; for CIFAR10, $\mathcal{L}_{rec}$ is more important.

\begin{table}[H]
	\centering
	\caption{}{OPTIMAL PARAMETERS FOR VARYING HASH CODES ACROSS THREE DATASETS}
	\\ [0.2cm]
	\setlength{\tabcolsep}{2.0mm}{
		\renewcommand{\arraystretch}{1.5}
		
		\begin{tabular}{c|cc|cc|cc}
			\hline
			\multirow{2}{*}{\textbf{Hash code}} & \multicolumn{2}{c|}{\textbf{AWA2}} & \multicolumn{2}{c|}{\textbf{CUB}} & \multicolumn{2}{c}{\textbf{CIFAR10}} \\ \cline{2-7} 
			& \multicolumn{1}{c|}{$\alpha$}    & $\beta$      & \multicolumn{1}{c|}{$\alpha$}     & $\beta$    & \multicolumn{1}{c|}{$\alpha$}       & $\beta$     \\ \hline
			24 bits                             & \multicolumn{1}{c|}{1.000}    & 1.000     & \multicolumn{1}{c|}{1.000}    & 0.500   & \multicolumn{1}{c|}{0.050}    & 1.000     \\
			48 bits                            & \multicolumn{1}{c|}{0.500}  & 0.001  & \multicolumn{1}{c|}{1.000}    & 0.001   & \multicolumn{1}{c|}{0.100}     & 0.100   \\
			64 bits                             & \multicolumn{1}{c|}{1.000}    & 0.010    & \multicolumn{1}{c|}{1.000}    & 1.000   & \multicolumn{1}{c|}{0.050}    & 1.000     \\
			128 bits                            & \multicolumn{1}{c|}{1.000}    & 0.010   & \multicolumn{1}{c|}{1.000}    & 1.000   & \multicolumn{1}{c|}{0.001}     & 0.100   \\ \hline
	\end{tabular}}
	\vspace{-0.2cm}
\end{table}

\emph{4) Visualization}: As shown in Fig. 9, classes (including both seen and unseen) are selected to present t-SNE visualizations of the hash codes acquired by RAZH fir the experimental datasets. The hash codes learned by RAZH have clearer distinguishing structures for these datasets. Taking the visualization results produced on AWA2 as examples, the hash codes of different classes of images learned by the method developed in this paper are effectively separated. The visualization results produced on the CUB dataset are loose because of the extensive number of classes and the limited number of instances within each class.

The clustering results obtained for the image patches are shown in Fig. 10. In the experiment, the image patches are divided into 9 clusters, with most of the patches belonging to the same attribute successfully clustered together. This indicates that the proposed clustering method is effective at gathering image patches that belong to the same attribute. The effectiveness of the RAZH alignment method is further validated by randomly selecting samples from the unseen classes in the AWA2 dataset for visualization purposes. As shown in Fig. 11, the initial column displays the original image, the second column shows the heatmap of the attention weights used by the model, and the third column shows the attribute labels corresponding to the mapped patches, where black represents the correct attributes and red represents the incorrect attributes. On the basis of the mapping attributes, RAZH provides accurate attribute correspondences for most patch mappings. Furthermore, the model heatmap shows that RAZH can effectively focus on the attribute regions of the target object, providing finer-grained attention.

\section{CONCLUSION}
In this paper, we propose a zero-shot hashing method based on reconstruction with part alignment; this approach aligns attributes and image patches via a reconstruction strategy to more effectively encode unseen classes. Compared with the existing zero-shot hashing methods, this approach aligns local image attributes with attribute semantics rather than aligning attributes with the entire image. In the hash learning stage, the model is jointly constrained by hash, classification, and reconstruction losses, effectively maintaining high similarity while mitigating overfitting to the seen classes and enhancing the ability to discriminate unseen-class hash codes. We extensively experiment on zero-shot hashing datasets, and RAZH demonstrates superior performance to that of many state-of-the-art methods in terms of various evaluation metrics.

Although RAZH demonstrates outstanding retrieval performance among the tested zero-shot hashing algorithms, there is still room for improvement. First, in terms of the number of model parameters, techniques such as model distillation can be introduced to further compress the model and achieve lightweight deployment. Second, while the model exhibits a fast inference procedure during the testing phase, the training phase is time-consuming because of the reconstruction process. To address this issue, more efficient optimization algorithms, such as adaptive optimization methods or distributed training, could be explored in the future to significantly reduce the required training time. Therefore, future work will focus on optimizing the model size and improving its training efficiency in these two areas.

\bibliographystyle{IEEEtran}
\bibliography{reference}

\vfill

\end{document}